\documentclass{article} 
\usepackage{iclr2027_conference,times}

\usepackage{amsmath,amsfonts,bm}

\def\eqref#1{equation~\ref{#1}}

\def\1{\bm{1}}

\DeclareMathAlphabet{\mathsfit}{\encodingdefault}{\sfdefault}{m}{sl}
\SetMathAlphabet{\mathsfit}{bold}{\encodingdefault}{\sfdefault}{bx}{n}

\usepackage{graphicx}
\usepackage{amssymb}
\usepackage{pifont}
\usepackage{array}
\usepackage{booktabs}
\usepackage{longtable}
\usepackage{float}
\usepackage{xcolor}
\usepackage{hyperref}
\hypersetup{hidelinks}
\usepackage{url}

\title{Event-STVG: Spatio-Temporal Video Grounding for Long-Form Event Reasoning}

\author{%
  {\bfseries
    Yizhe Zhang$^{1,*}$ \quad
    Xujia Wang$^{1,*}$ \quad
    Songqi Tang$^{1}$ \quad
    Xujie Tang$^{1}$  \quad
    Leo Jiahao Jiang$^{1}$
  } \\[0.2em]
  {\bfseries
    Jiacheng Wang$^{1}$ \quad
    Bowen Qu$^{2}$ \quad
    JinBo Li$^{2}$ \quad
    Qin Wang$^{2}$ \quad
    Ji Qi$^{1,\dagger}$ \quad
    Lei Hou$^{1}$
  } \\[0.2em]
  {\bfseries
    Bin Xu$^{1}$ \quad
    Juanzi Li$^{1}$
  } \\[0.6em]
  {\normalfont $^{1}$Tsinghua University} \\
  {\normalfont $^{2}$Beijing University of Posts and Telecommunications} \\[0.4em]
  {\normalfont\small
    $^{*}$Equal contribution. \quad
    $^{\dagger}$Correspondence: \texttt{qijimrc@gmail.com}
  }
}

\iclrfinalcopy

\begin{document}

\maketitle
\lhead{Preprint}

\begin{abstract}
Spatio-Temporal Video Grounding (STVG) aims to jointly localize a target in space and time given a natural-language query. Despite continued advances, existing STVG benchmarks remain largely centered on localizing targets through instance-specific descriptions, leaving them largely decoupled from long-form reasoning increasingly emphasized in multimodal large language models (MLLMs). In this paper, we introduce Event-STVG, a task that shifts STVG from locating what is described toward event-level reasoning by unifying long-form context, repeated-event disambiguation, event localization, role inference, and multi-target grounding. To facilitate research on Event-STVG, we construct a large-scale benchmark comprising 4,036 annotated event occurrences across eight domains. Experiments with state-of-the-art MLLMs reveal substantial difficulty, with the strongest model achieving only 5.14\% m-vIoU at the 10-minute horizon. Across models, we observe declining target-hit rates in longer contexts, divergent temporal and spatial grounding capabilities, and persistent difficulties in semantic-role identification and complete multi-role grounding. Event-STVG establishes a new challenge for visual understanding in MLLMs, calling for future models that can jointly reason about complex events across space and time.
\end{abstract}

\begin{figure}[h]
    \centering
    \includegraphics[width=\linewidth]{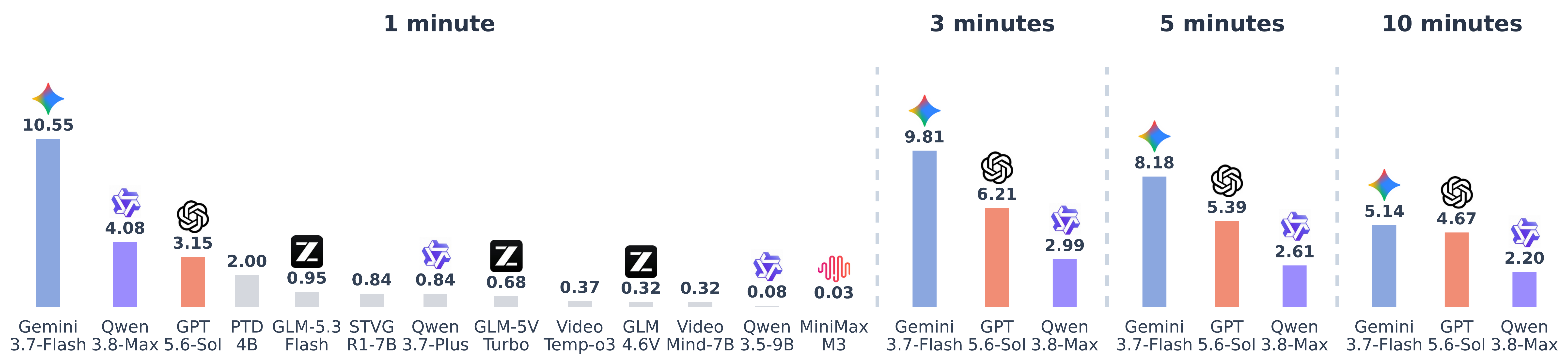}
    \caption{Model performance on Event-STVG across different video durations.}
    \label{fig:model-performance}
\end{figure}

\begin{figure}[t]
    \centering
    \includegraphics[width=\linewidth,pagebox=cropbox]{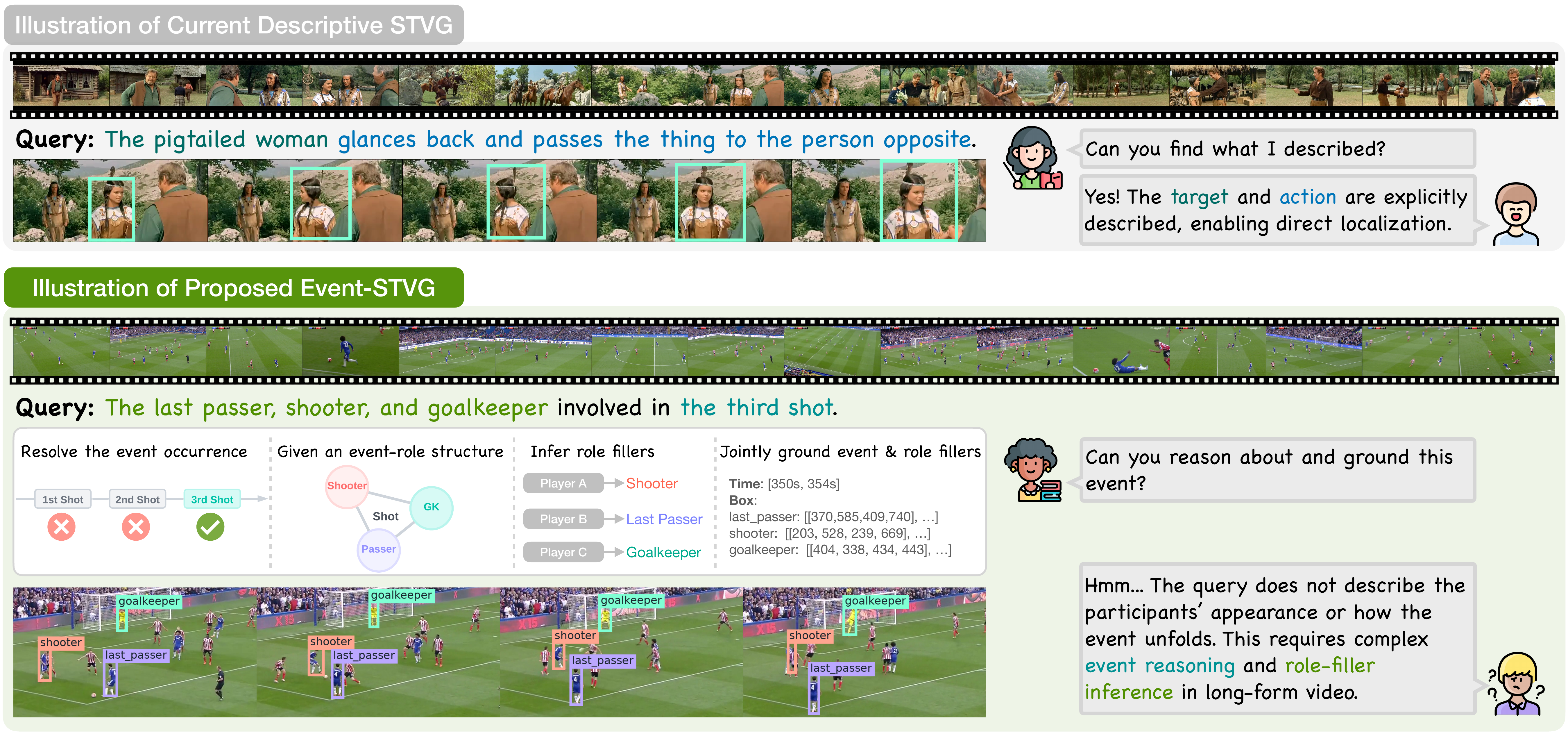}
    \caption{Comparison between Event-STVG and existing STVG. Existing STVG primarily performs description-guided localization, whereas Event-STVG requires complex event-level reasoning.}
    \label{fig:fig1}
\end{figure}

\section{Introduction}

Spatio-Temporal Video Grounding (STVG) aims to jointly localize a target of interest in space and time within an untrimmed video given a natural-language query~\citep{zhang2020vidstg}. By linking natural-language queries to precise spatio-temporal localization in video, STVG provides a concrete means of evaluating fine-grained multimodal grounding. Research in this area has advanced both grounding methods~\citep{yang2022tubedetr,rasheed2026ptd} and task formulations, with recent work extending STVG to multiple targets~\citep{yao2026omnistvg}, long-form videos~\citep{gu2026lfstvg}, and compositional queries~\citep{wang2026compstvg}.

Despite their growing complexity, existing STVG tasks remain largely centered on description-guided localization. For example, the query in Figure~\ref{fig:fig1} identifies a woman through her appearance, action, and relation to another person. As a result, current STVG benchmarks primarily test whether models can match a detailed linguistic description to the corresponding visual instance. This formulation is well suited to precise referent localization, but places limited demands on the richer long-form reasoning capabilities increasingly expected of MLLMs.

To move STVG beyond description-guided localization, we introduce \textbf{Event-STVG}, a new formulation that turns spatio-temporal grounding into a test of long-form event reasoning in MLLMs. Rather than describing the target instance directly, Event-STVG specifies an event category, its ordinal occurrence, and a set of semantic roles, requiring models to resolve the target occurrence, infer the corresponding role fillers, and jointly ground the event interval and its participants. This brings STVG substantially closer to the long-context temporal reasoning, repeated-event disambiguation, structured role inference, and multi-entity grounding increasingly emphasized in recent MLLM benchmarks~\citep{cheng2026vstar,zou2026comet,wang2026stevidence,aparcedo2026vista,asgarov2026nest}. To study this formulation at scale, we construct \textbf{Event-STVG}, comprising 4,036 annotated event occurrences across 8 domains, 45 event types, and 27 semantic roles. Each event is organized into an Oracle event window and four extended contexts of 1, 3, 5, and 10 minutes, constructed from its source video. We evaluate all 4,036 events under the Oracle and 1-minute settings, while a domain-stratified subset of 820 events is evaluated at 3, 5, and 10 minutes to examine the impact of increasing temporal context. Comprehensive experiments with frontier MLLMs, two specialized STVG methods, and two agentic video reasoning methods reveal substantial challenges in joint event and role grounding. The strongest model achieves only 5.14\% m-vIoU with 10-minute context. More importantly, our diagnostic analyses reveal that longer contexts substantially reduce target-hit rates and increase confusion among repeated same-type events. The decomposition of m-vIoU further reveals contrasting temporal and joint-spatial grounding strengths across models, indicating that stronger temporal localization does not necessarily translate into better role grounding. Moreover, semantic-role identification remains difficult even with candidate entities provided, while complete multi-role grounding is rare even under Oracle event windows. \textbf{These findings expose a persistent gap between resolving when an event occurs and grounding the structured participants that constitute it.}

Our main contributions are threefold:
\par\noindent (1) We rethink what STVG should evaluate for MLLMs and introduce \textbf{Event-STVG}, a task reframing STVG from description-guided localization toward a test of long-form event reasoning.
\par\noindent (2) We construct \textbf{Event-STVG}, a benchmark of 4,036 annotated events across 8 domains, and evaluate frontier MLLMs, specialized STVG models, and agentic methods, revealing substantial challenges in joint spatio-temporal grounding.
\par\noindent (3) We conduct \textbf{fine-grained diagnostic analyses} by decomposing m-vIoU. Together with analyses of repeated-event confusion, semantic-role identification, and complete multi-role grounding, we reveal \textbf{distinct model capabilities and failure modes} beyond aggregate scores.

\section{Related Work}
\label{sec:related-work}

\definecolor{benchgreen}{HTML}{16845B}
\definecolor{benchred}{HTML}{B7474D}
\newcommand{\benchyes}{\textcolor{benchgreen}{\ding{51}}}
\newcommand{\benchno}{\textcolor{benchred}{\ding{55}}}
\newcommand{\benchours}{\textbf{Event-STVG}}

\begin{table}[H]
\caption{Comparison with related benchmarks. Sizes: reported scale across all settings; Duration: mean video length. TG/SG: temporal/spatial localization; Joint: joint spatio-temporal localization; Ordinal Event–Role Grounding: occurrence-based event selection with joint temporal and multi-role trajectory grounding. M: manual; A\&M: automatic and manual; NR: not reported. Event-STVG includes 4,036 targets across four full STVG settings (1/3/5/10 min, 16,144 instances, mean 285\,s).}
\label{tab:benchmark-comparison}
\centering
\setlength{\tabcolsep}{2pt}
\renewcommand{\arraystretch}{1.0}
\begin{tabular*}{\textwidth}{@{\extracolsep{\fill}}l r r c c c c c@{}}
\toprule
\textbf{Benchmark} & \multicolumn{1}{r}{\textbf{Sizes}} & \multicolumn{1}{c}{\textbf{Duration (s)}} & \textbf{Anno} & \textbf{TG} & \textbf{SG} & \textbf{Joint} & \shortstack{\textbf{Ordinal}\\\textbf{Event-Role}\\\textbf{Grounding}} \\
\midrule
VideoZeroBench & 138 & 667.14 & M & \benchyes & \benchyes & \benchno & \benchno \\
CaST-Bench & 1,015 & 13.68 & A\&M & \benchyes & \benchyes & \benchyes & \benchno \\
ST-Evidence & 2,896 & NR & A\&M & \benchyes & \benchyes & \benchyes & \benchno \\
VidEvent & 1,110 & 82 & M & \benchyes & \benchno & \benchno & \benchno \\
NEST & 1,005 & 5,872.2 & A\&M & \benchyes & \benchno & \benchno & \benchno \\
MultiVENT-G & 1,168 & 83.7 & M & \benchyes & \benchyes & \benchno & \benchno \\
MultiSports & 3,200 & 20.9 & A\&M & \benchyes & \benchyes & \benchyes & \benchno \\
SportsHHI & 160 & 24.1 & M & \benchno & \benchyes & \benchno & \benchno \\
GroundingYouTube & 250 & NR & A\&M & \benchyes & \benchyes & \benchyes & \benchno \\
VidSTG & 6,924 & 28.01 & M & \benchyes & \benchyes & \benchyes & \benchno \\
HC-STVG-v1 & 5,660 & 20 & A\&M & \benchyes & \benchyes & \benchyes & \benchno \\
BOSTVG & 10,018 & 36.5 & M & \benchyes & \benchyes & \benchyes & \benchno \\
LF-STVG & 2,000 $\times$ 3 & 180 & M & \benchyes & \benchyes & \benchyes & \benchno \\
STVG-CompBench & 2,000 & NR & A\&M & \benchyes & \benchyes & \benchyes & \benchno \\
\midrule
\benchours & \textbf{4,036 $\times$ 4} & \textbf{285} & M & \benchyes & \benchyes & \benchyes & \benchyes \\
\bottomrule
\end{tabular*}
\end{table}

\paragraph{Spatio-Temporal Video Grounding}
Early STVG benchmarks ground language-referred entities in short videos~\citep{zhang2020vidstg,tang2022hcstvg}, spurring continued methodological advances~\citep{yang2022tubedetr,jin2022stcat,li2025llavast,zhang2026stvgr1,gao2026sarlstg,rasheed2026ptd}. Meanwhile, benchmark extensions to STVG address multiple targets~\citep{ji2024dvdst,yao2026omnistvg}, compositional queries~\citep{wang2026compstvg}, specialized domains~\citep{otsubo2026anygroundbench,xu2025togbench}, and minute-scale contexts~\citep{gu2026lfstvg}. Yet these formulations remain largely description-guided, with instance-specific appearance, action, or relational cues specifying what should be grounded. \textbf{Existing STVG therefore primarily evaluates description-conditioned localization rather than more complex event-level reasoning in MLLMs.}

\paragraph{Structured and Grounded Event Understanding}
Event understanding has long been studied through predicates, semantic arguments, and visual grounding, evolving from images~\citep{yatskar2016situation,pratt2020grounded} to short videos~\citep{shang2017vidvrd,xiao2020vrgv,sadhu2021visual,li2021multisports,khan2022grounded,wu2024sportshhi,darur2026identity}. Recent research has extended to minute-scale and longer videos, exploring temporal event localization without role trajectories~\citep{liang2025videvent,asgarov2026nest}, temporal and spatial grounding through distinct queries or stages~\citep{yang2023rest,sanders2024grounding}, and joint spatio-temporal grounding of action regions~\citep{chen2024groundingyoutube}. \textbf{To our knowledge, no prior benchmark requires models to reason over repeated event occurrences and jointly localize a selected event and its multiple semantic-role trajectories in minute-scale video contexts.}

\paragraph{Video QA with Spatio-Temporal Evidence}
General video QA benchmarks primarily evaluate answer correctness~\citep{fu2025videomme,zhou2025mlvu,wang2025lvbench}, while recent benchmarks increasingly require visual evidence through temporal localization~\citep{chen2025cgbench,huang2026egvqa,zou2026comet,liu2026trace} or decomposed spatio-temporal grounding~\citep{deng2025groundmore,cheng2026vstar,meng2026videozerobench}. More recently, E-VQA jointly grounds objects or regions supporting an answer~\citep{wang2026stevidence}, while CaST-Bench localizes evidence along a causal reasoning chain~\citep{zhang2026castbench}. \textbf{These benchmarks ground evidence supporting an answer or reasoning process, rather than jointly localizing an event and the semantic-role fillers that constitute its structure.}

\section{Event-STVG}

\subsection{Task Formulation}
Distinct from prior formulations, Event-STVG jointly requires temporal localization of the specified event occurrence and spatial grounding of its constituent roles, testing event–role reasoning in long-form video. The task can be formulated as follows. Given a long-form video \(V\), an event type \(E\) may occur repeatedly, forming an ordered sequence of event instances

\[
\mathcal{E}(V,E)=\langle E_1,E_2,\ldots,E_n\rangle.
\]

Each event type is associated with a semantic role schema

\[
\mathcal{R}(E)=\{r_1,r_2,\ldots,r_K\},
\]

which specifies the constituent roles involved in the event. Given an event type \(E\), an ordinal index \(m\), and a set of required roles \(\mathcal{R}_q\subseteq\mathcal{R}(E)\), Event-STVG asks a model to identify the \(m\)-th occurrence \(E_m\) and jointly return its temporal interval

\[
\tau_m=[t_m^s,t_m^e]
\]

together with the spatial trajectory of each queried role within that interval,

\[
\mathcal{B}_m=\{\mathbf{B}_{m,r}\mid r\in\mathcal{R}_q\},
\]

where \(\mathbf{B}_{m,r}\) denotes the sequence of bounding boxes grounding the entity that fills role \(r\). Thus, the query specifies \textbf{what event}, \textbf{which occurrence}, and \textbf{which roles}, but provides no instance-specific description of the participants, their appearance, or how the event unfolds.

To reduce ambiguity in annotation and evaluation, we define event-specific temporal boundaries using canonical role actions, specifying only when an event begins and ends rather than how its constituent roles interact. For example, a soccer shot may be defined from possession by the last passer to the point at which the shot outcome becomes clear. Different occurrences may nevertheless realize the same event structure differently, such as through a cross followed by a header or a through ball followed by a shot. These instance-specific interaction patterns are neither encoded in the query nor prescribed by the event definition, and must therefore be inferred from the video to resolve the target occurrence and ground its constituent roles.

\begin{figure}[t]
    \centering
    \includegraphics[width=\linewidth,pagebox=cropbox]{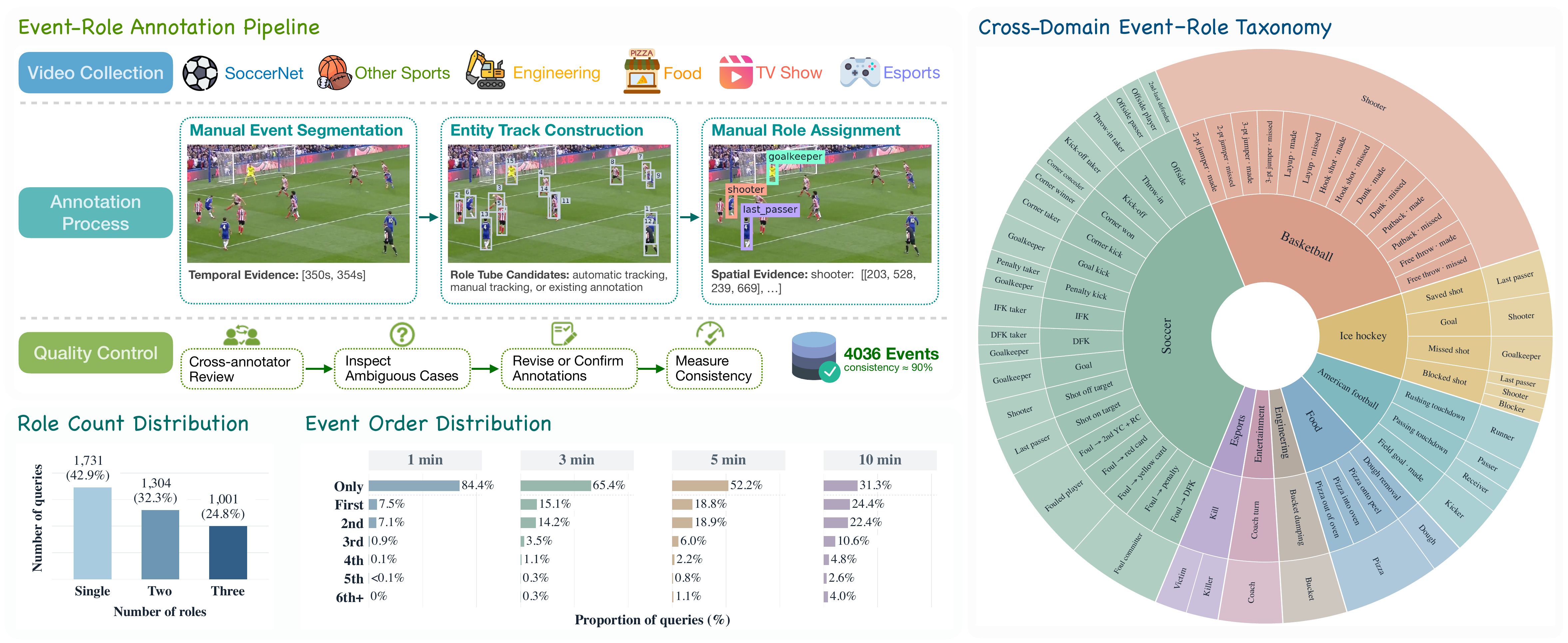}
    \caption{Overview of Event-STVG benchmark construction and composition. We construct Event-STVG through carefully designed human annotation with automated tracking assistance and multi-stage quality control. The resulting benchmark exhibits substantial diversity in event types and semantic roles, covering major soccer events and seven additional domains.}
    \label{fig:dataset-overview}
\end{figure}

\subsection{Benchmark Construction}

\textbf{Data Sources and Scope} Under this formulation, Event-STVG naturally applies to diverse real-world scenarios in which events recur over time with shared role structures. Examples include grounding the participants in a particular restaurant order, the entities in a specific traffic accident, or the part produced in a given manufacturing cycle. Considering both the availability of existing video datasets and our emphasis on structured multi-participant activities, we instantiate our benchmark primarily with soccer, complemented by basketball, ice hockey, football, engineering, food preparation, entertainment programs, and esports as shown in Figure~\ref{fig:dataset-overview}. The videos are drawn from SoccerNet-v2~\citep{deliege2021soccernetv2} and publicly available YouTube videos.

\textbf{Human Annotation} As illustrated in Figure~\ref{fig:dataset-overview}, annotation comprises three core components across all domains: event segmentation, entity track construction, and semantic-role assignment. For the four sports domains, we retrieve candidate events from SoccerNet, ESPN NCAA and NFL play-by-play records, and official NHL event records. For the remaining domains, annotators manually identify event occurrences by reviewing source videos. Annotators then verify each target event, determine its temporal boundaries according to event-specific definitions, and assign semantic roles. Role trajectories are annotated on frames sampled at 2 FPS. For soccer, we use SoccerMaster~\citep{yang2026soccermaster} with tracking and re-identification tools to propose candidate boxes and identities, which annotators manually verify and refine. For other domains, annotators draw role boxes manually. All final event boundaries, role assignments, and trajectories are human-verified. In this way, we construct Oracle event windows and extended contexts of 1, 3, 5, and 10 minutes from the untrimmed source videos. All qualifying same-type event occurrences within each context are chronologically enumerated to determine target ordinal indices, including contextual occurrences without full spatial annotations. Event-STVG queries are then generated from the event type, ordinal index, and required semantic roles. 

\textbf{Quality Control} We assess annotation consistency on approximately 10\% of the data for both temporal segmentation and role grounding. The paired annotations achieve a mean t-IoU of 0.8974 and a frame-level micro vIoU of 0.9016, with missing boxes counted as zero. We further inspect approximately 5\% of soccer samples and 10\% of samples from the three stop-clock sports to verify ordinal indices and generated queries, with all the inspected annotations passing the checks. Appendix~\ref{app:dataset-construction} provides further details on annotation and quality control.

\subsection{Benchmark Statistics}

As shown in Figure~\ref{fig:dataset-overview}, our benchmark comprises \textbf{4,036 annotated event instances} across soccer and seven additional domains. Together, they cover \textbf{45 event types} and \textbf{27 semantic role labels}, with \textbf{7,342 event--role pairs} and \textbf{58,915 annotated keyframe bounding boxes}, supporting event occurrence selection and semantic role grounding. Notably, \textbf{57.11\%} of all event instances involve multiple semantic roles, making the joint grounding of multiple role fillers a common requirement. Ordinal disambiguation is also central to Event-STVG. Across the same 4,036 target queries, the proportion of context windows containing multiple qualifying occurrences of the queried event type increases from \textbf{15.63\%} at one minute to \textbf{68.68\%} at ten minutes.

\section{Experiment}

\subsection{Settings}
\label{sec:experiment-settings}

\paragraph{Evaluated Models}
We evaluate thirteen models and methods: Gemini-3.7 Flash~\citep{deepmind2026gemini37flash}, GPT-5.6 Sol~\citep{openai2026gpt56systemcard}, Qwen-3.8 Max~\citep{qwen2026qwen38}, Qwen-3.7 Plus~\citep{qwen2026qwen37plus}, Qwen-3.5 9B~\citep{qwen2026qwen35}, GLM-5.3 Flash~\citep{zai2026glm53flash}, GLM-5V Turbo~\citep{glmVTeam2026glm5vTurbo}, GLM-4.6V~\citep{glmVTeam2025glm45v}, MiniMax-M3~\citep{lai2026minimaxsparseattention}, PTD~\citep{rasheed2026ptd}, STVG-R1~\citep{zhang2026stvgr1}, VideoTemp-o3~\citep{liu2026videotemp}, and VideoMind-7B~\citep{liu2026videomind}. This selection spans leading proprietary models, open-weight models, locally deployable models, task-specialized STVG models, and agentic reasoning methods.

\paragraph{Evaluation Protocol}
We evaluate all 4,036 events under the oracle and 1-minute settings. Under the oracle setting, models are given the ground-truth event interval and evaluated only on role grounding, isolating spatial grounding from temporal retrieval. These oracle windows are therefore excluded from the benchmark size reported in Table~\ref{tab:benchmark-comparison}. Building on the 1-minute setting, which provides the primary short-context evaluation of end-to-end Event-STVG, we further evaluate Gemini-3.7 Flash, GPT-5.6 Sol, and Qwen-3.8 Max on a fixed stratified subset of 820 events under each of the 3-, 5-, and 10-minute settings to assess long-context robustness. Prompt templates and model evaluation details are provided in Appendices~\ref{app:prompt-templates} and~\ref{app:evaluation-protocol}, respectively.

\paragraph{Metrics}
Temporal IoU measures the intersection-over-union of predicted and ground-truth event intervals~\citep{zhang2020vidstg,tang2022hcstvg,yang2022tubedetr}. For event \(i\) with \(R_i\) queried roles, we extend single-target vIoU~\citep{zhang2020vidstg,tang2022hcstvg} as
\begin{equation}
\mathrm{vIoU}_{i,r} =
\frac{\sum_{t\in S_{i,r}^{I}}\mathrm{IoU}(\hat{b}_{i,r}^{t},b_{i,r}^{t})}{|S_{i,r}^{U}|},
\qquad
\mathrm{vIoU}_i = \frac{1}{R_i}\sum_{r=1}^{R_i}\mathrm{vIoU}_{i,r}.
\label{eq:spatiotemporal-metrics}
\end{equation}
Here \(\hat b\) and \(b\) are predicted and ground-truth boxes, and \(S^I\) and \(S^U\) are the intersection and union of their sampled-frame supports. We report m-tIoU and m-vIoU by averaging temporal IoU and event-level vIoU equally over events. Frames absent from both tubes are excluded, and missing or invalid predictions contribute zero. The \(\mathrm{vIoU}@0.3\) and \(\mathrm{vIoU}@0.5\) scores are the proportions of events whose vIoU exceeds the respective thresholds.

\subsection{Main Results}

\begin{table}[!htbp]
\setlength{\belowcaptionskip}{\baselineskip}
\caption{Performance on Event-STVG (\%) under the Oracle and 1-minute settings. \textbf{Bold} indicates the best result in each column. VideoMind and VideoTemp-o3 use a spatial backend.}
\label{tab:main-oracle-one-minute}
\centering
\setlength{\tabcolsep}{3pt}
\renewcommand{\arraystretch}{1.12}
\begin{tabular*}{\textwidth}{@{\extracolsep{\fill}}lrrrrrrr@{}}
\toprule
& \multicolumn{3}{c}{Oracle} & \multicolumn{4}{c}{1 minute} \\
\cmidrule(lr){2-4}\cmidrule(l){5-8}
Model & m-vIoU & vIoU@0.3 & vIoU@0.5 & m-tIoU & m-vIoU & vIoU@0.3 & vIoU@0.5 \\
\midrule
\multicolumn{8}{@{}l@{}}{\textbf{\textit{General-Purpose VLMs}}} \\
\midrule
Gemini-3.7 Flash & \textbf{25.70} & \textbf{40.19} & \textbf{14.67} & 44.13 & \textbf{10.55} & \textbf{12.81} & \textbf{3.77} \\
GPT-5.6 Sol & 15.03 & 16.90 & 6.76 & \textbf{48.80} & 3.15 & 1.68 & 0.55 \\
Qwen-3.8 Max & 11.50 & 12.66 & 3.34 & 45.85 & 4.08 & 1.91 & 0.27 \\
Qwen-3.7 Plus & 3.15 & 1.34 & 0.37 & 38.10 & 0.84 & 0.35 & 0.05 \\
Qwen-3.5 9B & 3.29 & 2.95 & 0.82 & 14.16 & 0.08 & 0.00 & 0.00 \\
GLM-5.3 Flash & 12.81 & 14.49 & 2.75 & 31.25 & 0.95 & 0.15 & 0.02 \\
GLM-5V Turbo & 9.61 & 8.85 & 3.12 & 17.01 & 0.68 & 0.27 & 0.05 \\
GLM-4.6V & 12.93 & 13.90 & 4.01 & 7.27 & 0.32 & 0.20 & 0.05 \\
MiniMax-M3 & 0.85 & 0.47 & 0.07 & 3.98 & 0.03 & 0.00 & 0.00 \\
\midrule
\multicolumn{8}{@{}l@{}}{\textbf{\textit{STVG-Specialized VLMs}}} \\
\midrule
PTD-4B & 14.41 & 14.44 & 3.42 & 15.83 & 2.00 & 0.57 & 0.15 \\
STVG-R1-7B & 15.49 & 21.29 & 12.87 & 8.82 & 0.84 & 0.25 & 0.25 \\
\midrule
\multicolumn{8}{@{}l@{}}{\textbf{\textit{Agentic Video Temporal Grounding Methods}}} \\
\midrule
VideoMind-7B & 6.68 & 9.41 & 3.71 & 11.61 & 0.32 & 0.25 & 0.00 \\
VideoTemp-o3 & 6.68 & 9.41 & 3.71 & 11.79 & 0.37 & 0.25 & 0.00 \\
\bottomrule
\end{tabular*}
\end{table}

\paragraph{Short-context Event-STVG is already challenging for current models.}
Table~\ref{tab:main-oracle-one-minute} compares the thirteen models and methods. Even with the ground-truth event interval, the best Oracle m-vIoU is only 25.70\%, highlighting the difficulty of grounding event roles. Performance drops further in the 1-minute setting, where nine of the thirteen methods obtain m-vIoU below 1\%. The STVG specialists PTD-4B and STVG-R1-7B achieve only 2.00\% and 0.84\% m-vIoU, respectively, while VideoMind-7B and VideoTemp-o3 achieve 0.32\% and 0.37\% with a spatial backend. Model strengths also differ across components: GPT achieves the highest temporal localization score at 1 minute, whereas Gemini achieves the highest joint grounding score.

\begin{table}[!htbp]
\setlength{\belowcaptionskip}{\baselineskip}
\caption{One-minute m-vIoU (\%) across eight domains. Domains are ranked by mean m-vIoU across the thirteen models. \textbf{Bold} indicates the best result in each domain.}
\label{tab:multidomain-performance}
\centering
\small
\setlength{\tabcolsep}{0.8pt}
\renewcommand{\arraystretch}{1.12}
\makebox[\linewidth][c]{%
\begin{tabular*}{\linewidth}{@{\extracolsep{\fill}}lrrrrrrrrrrrrrr@{\hspace{20pt}}}
\toprule
Domain & Mean & \makebox[0pt][l]{\hspace{-20pt}\raisebox{\depth}{\rotatebox{60}{\strut Gemini-3.7 Flash}}} & \makebox[0pt][l]{\hspace{-20pt}\raisebox{\depth}{\rotatebox{60}{\strut GPT-5.6 Sol}}} & \makebox[0pt][l]{\hspace{-20pt}\raisebox{\depth}{\rotatebox{60}{\strut Qwen-3.8 Max}}} & \makebox[0pt][l]{\hspace{-20pt}\raisebox{\depth}{\rotatebox{60}{\strut Qwen-3.7 Plus}}} & \makebox[0pt][l]{\hspace{-20pt}\raisebox{\depth}{\rotatebox{60}{\strut Qwen-3.5 9B}}} & \makebox[0pt][l]{\hspace{-20pt}\raisebox{\depth}{\rotatebox{60}{\strut GLM-5.3 Flash}}} & \makebox[0pt][l]{\hspace{-20pt}\raisebox{\depth}{\rotatebox{60}{\strut GLM-5V Turbo}}} & \makebox[0pt][l]{\hspace{-20pt}\raisebox{\depth}{\rotatebox{60}{\strut GLM-4.6V}}} & \makebox[0pt][l]{\hspace{-20pt}\raisebox{\depth}{\rotatebox{60}{\strut MiniMax-M3}}} & \makebox[0pt][l]{\hspace{-20pt}\raisebox{\depth}{\rotatebox{60}{\strut PTD-4B}}} & \makebox[0pt][l]{\hspace{-20pt}\raisebox{\depth}{\rotatebox{60}{\strut STVG-R1-7B}}} & \makebox[0pt][l]{\hspace{-20pt}\raisebox{\depth}{\rotatebox{60}{\strut VideoMind-7B}}} & \makebox[0pt][l]{\hspace{-20pt}\raisebox{\depth}{\rotatebox{60}{\strut VideoTemp-o3}}} \\
\midrule
Entertainment & 14.04 & 35.62 & \textbf{42.11} & 23.55 & 19.98 & 1.09 & 5.68 & 7.06 & 9.15 & 0.66 & 8.18 & 19.05 & 10.40 & 0.00 \\
Engineering & 11.32 & 23.33 & \textbf{33.32} & 20.04 & 14.03 & 0.06 & 10.94 & 19.88 & 4.75 & 0.00 & 20.77 & 0.00 & 0.00 & 0.00 \\
Food & 5.80 & \textbf{23.57} & 19.65 & 15.24 & 6.27 & 0.00 & 4.70 & 1.61 & 2.99 & 0.03 & 0.82 & 0.00 & 0.45 & 0.10 \\
Football & 3.11 & \textbf{12.26} & 7.06 & 1.94 & 2.45 & 0.05 & 2.21 & 2.41 & 1.17 & 0.03 & 4.08 & 3.50 & 0.58 & 2.65 \\
Esports & 2.14 & \textbf{12.02} & 3.93 & 0.75 & 3.16 & 0.13 & 1.32 & 2.26 & 0.78 & 0.02 & 3.43 & 0.00 & 0.00 & 0.00 \\
Basketball & 1.82 & \textbf{7.06} & 4.18 & 2.58 & 1.83 & 0.07 & 1.86 & 1.03 & 1.23 & 0.06 & 3.37 & 0.27 & 0.13 & 0.00 \\
Soccer & 1.65 & \textbf{10.50} & 2.27 & 3.95 & 0.42 & 0.08 & 0.74 & 0.40 & 0.13 & 0.02 & 1.72 & 0.68 & 0.23 & 0.34 \\
Ice hockey & 0.85 & \textbf{3.65} & 2.83 & 1.47 & 0.45 & 0.06 & 0.39 & 0.14 & 0.11 & 0.04 & 0.82 & 0.42 & 0.35 & 0.31 \\
\bottomrule
\end{tabular*}%
}
\end{table}

\paragraph{Short-context Event-STVG remains challenging across domains.} Table 3 shows substantial variation across domains under the one-minute setting, with sports yielding lower mean scores than entertainment, engineering, and food. Averaged over the thirteen models, m-vIoU ranges from 0.85\% in ice hockey to 14.04\% in entertainment. Yet even the strongest model--domain combination, GPT in entertainment, achieves only 42.11\% m-vIoU, leaving substantial room for improvement.

\begin{table}[!htbp]
\setlength{\belowcaptionskip}{\baselineskip}
\caption{Performance on the same 820 Event-STVG targets (\%) under Oracle and 1-, 3-, 5-, and 10-minute settings. Oracle and 1-minute results are computed on this subset. \textbf{Bold} indicates the best result in each column.}
\label{tab:main-long-context}
\centering
\small
\setlength{\tabcolsep}{2pt}
\renewcommand{\arraystretch}{1.12}
\begin{tabular*}{\textwidth}{@{\extracolsep{\fill}}lrrrrrrrrr@{}}
\toprule
& \multicolumn{1}{c}{Oracle} & \multicolumn{2}{c}{1 minute} & \multicolumn{2}{c}{3 minutes} & \multicolumn{2}{c}{5 minutes} & \multicolumn{2}{c}{10 minutes} \\
\cmidrule(lr){2-2}\cmidrule(lr){3-4}\cmidrule(lr){5-6}\cmidrule(lr){7-8}\cmidrule(l){9-10}
Model & m-vIoU & m-tIoU & m-vIoU & m-tIoU & m-vIoU & m-tIoU & m-vIoU & m-tIoU & m-vIoU \\
\midrule
Gemini-3.7 Flash & \textbf{31.41} & 45.96 & \textbf{10.74} & 38.47 & \textbf{9.81} & 34.04 & \textbf{8.18} & 25.30 & \textbf{5.14} \\
GPT-5.6 Sol & 21.30 & \textbf{49.90} & 6.57 & \textbf{46.23} & 6.21 & \textbf{43.16} & 5.39 & \textbf{38.69} & 4.67 \\
Qwen-3.8 Max & 11.51 & 43.76 & 4.77 & 37.25 & 2.99 & 32.10 & 2.61 & 24.24 & 2.20 \\
\bottomrule
\end{tabular*}
\end{table}

\paragraph{Longer contexts further exacerbate the grounding difficulty.}
We focus the subsequent long-context evaluation on Gemini, GPT, and Qwen, the three strongest general-purpose models in 1-minute m-vIoU. The 820-target subset has a different domain distribution from the full benchmark, so direct comparisons with the full-set Oracle and 1-minute scores in Table~\ref{tab:main-oracle-one-minute} would conflate context length with sample composition. Table~\ref{tab:main-long-context} therefore reports Oracle and all four context durations on the same 820 targets. Within this matched set, both temporal and joint grounding performance decrease monotonically from 1 to 10 minutes for all three models. Gemini’s m-vIoU falls from 10.74\% at 1 minute to 5.14\% at 10 minutes, which remains the best result at that horizon. Even this best ten-minute score is notably lower than the 65.73–71.49\% m-vIoU achieved by individual human annotators on a 30-query subset under the same duration and input setting (Appendix~\ref{app:human-ten-minute}), suggesting that current model performance remains far from saturation.

\subsection{Diagnostic Analysis}

\label{sec:diagnostic-analysis}

Aggregate m-vIoU scores reveal the difficulty of Event-STVG but do not distinguish failures in identifying the requested event occurrence, determining its temporal boundaries, and grounding its semantic roles. Prior STVG studies have examined temporal and spatial grounding separately~\citep{yang2022tubedetr,wang2026compstvg}, while recent temporal grounding research has highlighted zero-overlap failures and interference from competing event occurrences~\citep{nie2026emground,chen2026dtpopsd}. These distinctions are particularly relevant to Event-STVG, where repeated occurrences of the same event type introduce plausible alternatives to the target. To better characterize model failures, we factorize the existing m-vIoU into three interpretable components, followed by analyses of event-instance confusion, complete multi-role grounding, and semantic-role identification.

\paragraph{Factorizing End-to-End Grounding}

Let $V_i$ denote the event-level vIoU for query $i$, and $T_i=|I_i|/|U_i|$ denote temporal IoU on the same evaluation sampling grid, where $I_i$ and $U_i$ are the intersection and union of the predicted and ground-truth temporal supports. For $N$ queries, we define

\[
\begin{aligned}
H &= \frac{1}{N}\sum_i \mathbf{1}[T_i>0],\\
T_H &= \frac{\sum_i T_i}{\sum_i\mathbf{1}[T_i>0]},\\
S_w &= \frac{\sum_i V_i}{\sum_i T_i}.
\end{aligned}
\]

The existing metric admits the exact factorization

\begin{equation}
\mathrm{m\text{-}vIoU}=H\cdot T_H\cdot S_w.
\label{eq:diagnostic-factorization}
\end{equation}

Here, $H$ measures target-hit rate, $T_H$ measures temporal alignment conditional on a hit, and $S_w$ measures temporally weighted joint-spatial grounding quality. Specifically, $S_w$ is the temporal-IoU-weighted average of $S_i=V_i/T_i$ over temporally overlapping queries, preserving the original multi-role vIoU definition. This factorization does not modify the benchmark metric or assume independent cognitive components.

\paragraph{Finding I: Long-Context Degradation Is Largely Reflected in Reduced Target Hits.}

\begin{figure}[t]
\centering
\includegraphics[width=\linewidth]{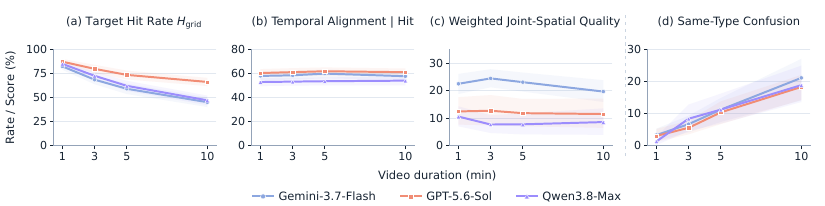}
\caption{Decomposing end-to-end grounding. (a--c) The same 820 matched queries, using grid $H$, $T_H$ and $S_w$ as defined in Eq.~\ref{eq:diagnostic-factorization}. (d) Same-type mismatches divided by 355 fixed semantically supported queries, not all 820. Mismatches exclude continuous semantic-GT-window target hits and use typed GT windows with audited aligned point fallback. Shading shows pointwise 95\% source-cluster intervals (20,000 resamples, seed 21).}
\label{fig:diagnostic-decomposition}
\end{figure}

Figure~\ref{fig:diagnostic-decomposition} presents the decomposition across video durations on the same 820 target queries. As context length increases, all three models exhibit a pronounced decline in target-hit rate, whereas temporal alignment conditional on a hit changes comparatively little. The weighted joint-spatial factor also declines in point estimate, but less markedly than the hit rate. These results suggest that long-context degradation is predominantly reflected in failures to overlap the requested event, rather than a comparable deterioration in temporal alignment among remaining hits. Nevertheless, temporal boundary quality is not entirely stable: among the 286 queries consistently hit by Gemini across all four durations, m-tIoU decreases by 3.84 percentage points from 1 to 10 minutes (95\% CI [$-$6.14, $-$1.55]; Holm-adjusted $p=0.0197$).

\paragraph{Confusion among repeated events}

A target miss does not necessarily indicate failure to recognize the event type; the model may instead localize another occurrence of the same type. We therefore examine misses overlapping a different qualifying same-type occurrence. As shown in Figure~\ref{fig:diagnostic-decomposition}, same-type mismatch rates increase significantly with context length for all three models on a fixed subset of 355 queries with sufficient occurrence-level temporal evidence. These mismatches account for a substantial proportion of long-context misses, although their prevalence varies across models and does not explain all failures (Appendix~\ref{app:same-type-misses}). This finding highlights the additional challenge of selecting the requested occurrence among repeated events, beyond recognizing the event category.

\paragraph{Finding II: Temporal Grounding Does Not Guarantee Joint-Spatial Grounding.}

The exact decomposition further reveals contrasting temporal and joint-spatial grounding profiles across models (Figure~\ref{fig:diagnostic-decomposition}). GPT consistently achieves higher target-hit rates and stronger conditional temporal alignment than Gemini. In contrast, Gemini achieves substantially higher temporally weighted joint-spatial quality, resulting in relatively similar end-to-end m-vIoU scores despite their different temporal performance. This difference persists when controlling for the set of temporally hit queries. On the same 249 queries hit by all three models at 10 minutes, Gemini exceeds GPT in weighted joint-spatial quality by 8.35 percentage points (95\% CI [5.01, 12.17]). The corresponding query-level normalized scores also differ significantly after multiple-comparison correction. Thus, stronger temporal grounding does not necessarily translate into better localization of event-specific semantic-role trajectories. This disparity illustrates why temporal localization alone is insufficient to characterize Event-STVG performance. 

\paragraph{Finding III: Complete Multi-Role Grounding Remains Challenging Even with Oracle Windows.}

Beyond average joint-spatial quality, Event-STVG requires models to localize all requested roles within the same event. We therefore examine event-level grounding completeness. For multi-role queries, predictions are classified as All, Some, or None, depending on whether all, a nonempty proper subset, or none of the required role trajectories achieve full-trajectory vIoU greater than 0.3.

\begin{figure}[t]
\centering
\includegraphics[width=\linewidth]{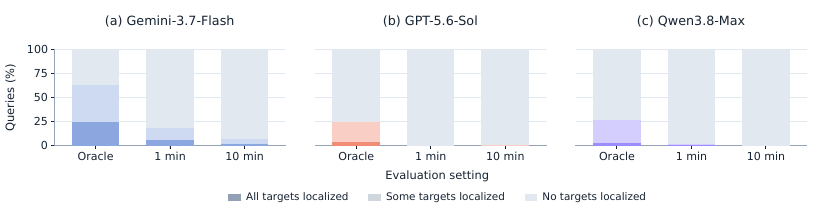}
\caption{Complete multi-role grounding on 392 matched queries (210 two-role, 182 three-role; 80 sources). Panels (a--c) show Gemini, GPT, and Qwen, respectively. Model-colored, lighter, and gray segments denote All, Some, and None of the required roles localized. Success requires per-role full-trajectory vIoU strictly greater than 0.3. Oracle, 1-minute, and 10-minute results are shown.}
\label{fig:complete-multirole}
\end{figure}

As shown in Figure~\ref{fig:complete-multirole}, complete multi-role grounding remains uncommon across models, even under Oracle temporal windows. Performance deteriorates further with longer contexts, where most predictions fail to localize any required role successfully. Notably, even the strongest model achieves only 25.26\% complete grounding under Oracle, falling to 2.04\% at 10 minutes on the same 392 multi-role queries. These results indicate that providing the correct event interval does not ensure successful grounding of all event participants. Moreover, the difference between partial and complete localization demonstrates that average role-level grounding quality alone does not fully characterize whether the required event structure has been recovered.

\begin{table}[t]
\centering
\small
\setlength{\tabcolsep}{4pt}
\caption{Semantic-role identification under Oracle candidate-ID selection. Results are percentages with 95\% match-clustered confidence intervals.}
\label{tab:oracle-id-accuracy}
\begin{tabular*}{\linewidth}{@{\extracolsep{\fill}}lrr@{}}
\toprule
Model & ID Accuracy & Joint ID Accuracy \\
\midrule
Gemini-3.7 Flash & 62.95 [58.87, 69.45] & 48.56 [44.20, 55.62] \\
GPT-5.6 Sol & 45.03 [42.29, 49.86] & 29.90 [25.37, 35.67] \\
Qwen3.8-Max & 47.26 [42.13, 55.71] & 33.77 [28.66, 41.98] \\
\bottomrule
\end{tabular*}
\end{table}

\paragraph{Semantic-Role Identification through Candidate-ID Selection}

Although vIoU is a standard metric for STVG, it conflates several sources of spatial error, including incorrect entity selection, imprecise bounding boxes, and incomplete trajectories. Consequently, the low joint-spatial quality and incomplete multi-role grounding observed above do not reveal whether models can identify the correct semantic-role fillers in the first place.

To examine this ability more directly, we introduce a complementary candidate-ID selection diagnostic under Oracle temporal windows. Rather than generating boxes freely, models select the displayed candidate ID corresponding to each queried semantic role at prescribed timestamps. By providing the event interval and candidate boxes, this setting removes temporal retrieval and free-form box generation from the required output. 

We evaluate a separate frozen cohort of 300 soccer queries, covering 17 event categories and 48 matches. ID Accuracy measures the proportion of GT-visible role--frame pairs for which the selected ID exactly matches the GT-associated candidate ID. GT-invisible pairs are excluded, while incorrect IDs, null, missing predictions, and invalid responses count as errors. We further report Joint ID Accuracy, requiring all GT-visible roles at a timestamp to be identified correctly. Both metrics pool eligible observations across the cohort, with confidence intervals estimated using match-clustered bootstrap. More details of the experiment can be found in Appendix~\ref{app:oracle-id-selection}.

As shown in Table~\ref{tab:oracle-id-accuracy}, Gemini achieves the strongest semantic-role identification performance, significantly outperforming GPT and Qwen, while still substantially lower than the 96.80\% human baseline illustrated in Appendix~\ref{app:human-id-selection}. Joint ID Accuracy is consistently lower than individual ID Accuracy, indicating that correctly identifying individual role fillers does not consistently translate into identifying the complete visible role set. Importantly, substantial identification errors remain even when the target event interval is provided and a correct candidate is available for every evaluated visible role. This demonstrates that semantic-role identification remains challenging even when temporal localization and free-form box generation are removed. Since candidates and identity associations are supplied, the results characterize candidate-assisted role selection rather than unrestricted visual tracking or an input-controlled comparison of model architectures.

Together, these analyses provide complementary views of Event-STVG performance: whether models reach the requested event, how accurately they localize it after a hit, how much temporal alignment translates into joint-spatial grounding, and whether they can identify the semantic-role fillers when candidate entities are supplied. These analyses show that Event-STVG is not merely challenging under the evaluated settings, but also provides a more fine-grained understanding of model behavior beyond aggregate m-vIoU scores.

\section{Conclusion}
In this paper, we introduce \textbf{Event-STVG}, a new formulation of spatio-temporal video grounding that shifts description-guided localization toward structured event reasoning. We construct a benchmark with 4,036 manually annotated event occurrences across eight domains, which requires models to resolve a target occurrence among repeated events, infer its role fillers, and jointly ground them in space and time. Our experiments show that current models struggle with Event-STVG, with the best achieving only 5.14\% m-vIoU at 10 minutes. Our analyses further reveal (1) declining target-hit rates and growing same-type confusion in longer contexts; (2) contrasting temporal and joint-spatial grounding strengths; and (3) persistent role-identification difficulties despite candidate assistance and rare complete multi-role grounding even under Oracle windows. Event-STVG thus contributes to the growing emphasis on complex reasoning in MLLM evaluation by grounding structured event understanding in verifiable spatio-temporal predictions.

\subsection*{AI Use Statement}
Generative AI tools were used to assist with literature research, manuscript drafting and editing, and code development and debugging. They were not used to generate or annotate benchmark ground truth. All AI-assisted materials were reviewed and verified by the authors, who take full responsibility for the final manuscript and results.

\subsection*{Reproducibility Statement}
The appendix provides detailed descriptions of benchmark construction, annotation and quality control, model prompting and input configurations, and evaluation metrics and protocols. We will publicly release the benchmark annotations and evaluation code to facilitate reproducibility.

\clearpage
\urlstyle{rm}
\bibliography{iclr2027_conference,related_work}
\bibliographystyle{iclr2027_conference}

\clearpage
\appendix
\section*{Appendices}
\section{Prompt Template and Output Format}
\label{app:prompt-templates}
\renewcommand{\thetable}{A.\arabic{table}}
\renewcommand{\theHtable}{A.\arabic{table}}
\setcounter{table}{0}
\renewcommand{\ttdefault}{cmtt}

We query each model with an event type, an ordinal occurrence, event-specific temporal boundaries, and the roles to ground. The following is the prompt template for a 10-minute soccer video and the \emph{Shot on Target} event. The ordinal placeholder is replaced by the requested occurrence index.

\begin{quote}
You will see a 10-minute soccer match video. Find the \{ordinal\} occurrence of Shot on Target. The event starts when the last passer first gains possession during the target attack, and ends when the outcome of the shot becomes clear. Draw bounding boxes for the last passer, the shooter, and the defending goalkeeper.

\medskip
\noindent\textbf{Event Definition:}

Shot on Target: An attacking sequence culminating in an intentional attempt to score that is directed toward the goal and either results in a goal or is prevented from entering the goal.

\medskip
\noindent\textbf{Role Definitions:}
\begin{itemize}
    \item[1)] \texttt{last\_passer}: The attacking player who makes the final effective pass to the shooter before the shot.
    \item[2)] \texttt{shooter}: The attacking player who takes the shot.
    \item[3)] \texttt{goalkeeper}: The goalkeeper defending the goal against the shot.
\end{itemize}

\medskip
\noindent\textbf{Instructions:}
\begin{itemize}
    \item[1)] Report the event's start and end times in seconds relative to the input video.
    \item[2)] Starting at the event's start time, provide role boxes every 0.5 seconds within the event window.
    \item[3)] For each sampled timestamp, include all three roles: \texttt{last\_passer}, \texttt{shooter}, and \texttt{goalkeeper}.
    \item[4)] Use tight bounding boxes in \texttt{[xmin, ymin, xmax, ymax]} format, normalized to $[0,1000]$, with the origin at the top-left of the frame.
    \item[5)] Track the same person for each role throughout the event. If that person is not reliably visible, output \texttt{null}; resume tracking the same person when visible again.
    \item[6)] Return only JSON with \texttt{temporal\_window} (\texttt{start\_sec}, \texttt{end\_sec}) and \texttt{role\_localization} (\texttt{time\_sec} and each role's box or \texttt{null}).
\end{itemize}
\end{quote}

\noindent\textbf{Note:} The event and role definitions provide general semantic guidance. They do not describe the target occurrence's specific action sequence or the appearance and concrete actions of its participants; these must be inferred from the video.

The required output is a JSON object with one temporal window and a list of role localizations. The example below illustrates the structure and coordinate format; its values are schematic.

\begin{verbatim}
{
  "temporal_window": {
    "start_sec": 123.0,
    "end_sec": 123.5
  },
  "role_localization": [
    {
      "time_sec": 123.0,
      "last_passer": [120, 240, 180, 500],
      "shooter": null,
      "goalkeeper": [810, 210, 860, 470]
    },
    {
      "time_sec": 123.5,
      "last_passer": [125, 240, 185, 500],
      "shooter": [420, 220, 475, 490],
      "goalkeeper": [805, 210, 855, 470]
    }
  ]
}
\end{verbatim}

\clearpage
\setcounter{section}{1}
\section{Evaluation Protocol and Implementation Details}
\label{app:evaluation-protocol}
\renewcommand{\thetable}{B.\arabic{table}}
\renewcommand{\theHtable}{B.\arabic{table}}
\setcounter{table}{0}

\subsection{Long-Context Cohort Construction}
\label{app:long-context-sampling}
Our goal is to compare how context duration affects event occurrence selection, temporal localization, and semantic-role grounding on a fixed set of targets while maintaining coverage of event categories. For the main long-context comparison of Gemini-3.7 Flash, GPT-5.6 Sol, and Qwen-3.8 Max, we select 300 targets from the 3,516-target soccer population and retain all 520 targets from the other seven domains. The resulting cohort contains 820 targets spanning all eight domains and 45 event categories. The same targets are used for Oracle and the 1-, 3-, 5-, and 10-minute comparisons.

For soccer, we allocate 300 slots approximately in proportion to the frequencies of the 17 event categories, using a constrained largest-remainder allocation with at least one target per category. Within these quotas, an integer program minimizes an equally weighted sum of absolute discrepancies in the joint distributions of event category and occurrence order at 3, 5, and 10 minutes, and in the joint distribution of event category and order trajectory across those durations. An order trajectory is the sequence of occurrence-order bins assigned to the same target at 3, 5, and 10 minutes. The selection bins are Only, First, 2nd, 3rd, and 4th+, where First refers to the first of multiple same-type occurrences. Targets are then sampled within the allocated cells defined by event category and order trajectory using the fixed seed 21.

Role count is not an additional selection criterion. Table~\ref{tab:long-context-role-composition} reports the observed proportions of queries requiring one, two, or three semantic roles after selection. Table~\ref{tab:long-context-ordinal-composition} compares the final cohort's exact occurrence-order distributions with the full benchmark, while Table~\ref{tab:soccer-sampling-ordinal-composition} compares the selected soccer component with its source population. All counts and percentages are computed from the fixed target membership and ground-truth annotations. The tables expand occurrence ranks through the largest observed rank at each duration, rather than retaining the coarse bins used during selection.

The final cohort contains 428 one-role queries (52.20\%), 210 two-role queries (25.61\%), and 182 three-role queries (22.20\%). The corresponding counts in the full benchmark are 1,731 (42.89\%), 1,304 (32.31\%), and 1,001 (24.80\%). Retaining all non-soccer targets changes the domain composition, so we report both the final cohort and the selected soccer component against their respective source populations.

\begin{table}[!ht]
\centering
\setlength{\tabcolsep}{4pt}
\renewcommand{\arraystretch}{1.12}
\caption{Distribution of the number of queried semantic roles. Entries report count (percentage within each cohort). Role counts are unchanged across context durations.}\label{tab:long-context-role-composition}
\begin{tabular*}{\linewidth}{@{\extracolsep{\fill}}lrrrr@{}}
\toprule
Cohort & Total & 1 role & 2 roles & 3 roles \\
\midrule
Full benchmark & 4,036 & 1,731 (42.89) & 1,304 (32.31) & 1,001 (24.80) \\
Long-context cohort & 820 & 428 (52.20) & 210 (25.61) & 182 (22.20) \\
\midrule
Full soccer & 3,516 & 1,424 (40.50) & 1,203 (34.22) & 889 (25.28) \\
Selected soccer & 300 & 121 (40.33) & 109 (36.33) & 70 (23.33) \\
\bottomrule
\end{tabular*}
\par\smallskip
\begin{minipage}{\linewidth}\textit{Notes.} The long-context cohort contains the selected soccer targets and all non-soccer targets. The lower two rows compare the sampled soccer component with its source population. Percentages describe the observed composition; they are rounded independently to two decimal places.\end{minipage}
\end{table}

\clearpage
\begingroup
\setlength{\tabcolsep}{4pt}
\renewcommand{\arraystretch}{1.05}
\setlength{\LTcapwidth}{\linewidth}
\newlength{\cohortordinalwidth}
\setlength{\cohortordinalwidth}{\dimexpr\linewidth-8\tabcolsep\relax}
\begin{longtable}{@{}>{\raggedright\arraybackslash}p{.30\cohortordinalwidth}*{4}{>{\raggedleft\arraybackslash}p{.175\cohortordinalwidth}}@{}}
\caption{Exact target occurrence-order distributions in the full benchmark and the long-context cohort. The same targets are used at 1, 3, 5, and 10 minutes within each cohort.}\label{tab:long-context-ordinal-composition}\\
\toprule
& \multicolumn{2}{c}{Full benchmark} & \multicolumn{2}{c}{Long-context cohort} \\
\cmidrule(lr){2-3}\cmidrule(l){4-5}
Target order & Count & Share (\%) & Count & Share (\%) \\
\midrule
\endfirsthead
\multicolumn{5}{@{}l}{\tablename\ \thetable\ (continued)}\\
\toprule
& \multicolumn{2}{c}{Full benchmark} & \multicolumn{2}{c}{Long-context cohort} \\
\cmidrule(lr){2-3}\cmidrule(l){4-5}
Target order & Count & Share (\%) & Count & Share (\%) \\
\midrule
\endhead
\midrule
\multicolumn{5}{r@{}}{Continued on next page}\\
\endfoot
\bottomrule
\endlastfoot
\multicolumn{5}{@{}l}{\textbf{1-minute context}}\\*
Only & 3,405 & 84.37 & 630 & 76.83 \\
First & 303 & 7.51 & 81 & 9.88 \\
2nd & 287 & 7.11 & 82 & 10.00 \\
3rd & 35 & 0.87 & 21 & 2.56 \\
4th & 5 & 0.12 & 5 & 0.61 \\
5th & 1 & 0.02 & 1 & 0.12 \\
\midrule
\textbf{Total} & \textbf{4,036} & \textbf{100.00} & \textbf{820} & \textbf{100.00} \\
\midrule
Repeated & 631 & 15.63 & 190 & 23.17 \\
\midrule
\multicolumn{5}{@{}l}{\textbf{3-minute context}}\\*
Only & 2,641 & 65.44 & 487 & 59.39 \\
First & 610 & 15.11 & 136 & 16.59 \\
2nd & 574 & 14.22 & 113 & 13.78 \\
3rd & 140 & 3.47 & 44 & 5.37 \\
4th & 46 & 1.14 & 21 & 2.56 \\
5th & 11 & 0.27 & 7 & 0.85 \\
6th & 6 & 0.15 & 5 & 0.61 \\
7th & 5 & 0.12 & 4 & 0.49 \\
8th & 3 & 0.07 & 3 & 0.37 \\
\midrule
\textbf{Total} & \textbf{4,036} & \textbf{100.00} & \textbf{820} & \textbf{100.00} \\
\midrule
Repeated & 1,395 & 34.56 & 333 & 40.61 \\
\midrule
\multicolumn{5}{@{}l}{\textbf{5-minute context}}\\*
Only & 2,105 & 52.16 & 404 & 49.27 \\
First & 760 & 18.83 & 153 & 18.66 \\
2nd & 762 & 18.88 & 122 & 14.88 \\
3rd & 242 & 6.00 & 64 & 7.80 \\
4th & 89 & 2.21 & 30 & 3.66 \\
5th & 33 & 0.82 & 15 & 1.83 \\
6th & 21 & 0.52 & 11 & 1.34 \\
7th & 9 & 0.22 & 6 & 0.73 \\
8th & 4 & 0.10 & 4 & 0.49 \\
9th & 3 & 0.07 & 3 & 0.37 \\
10th & 2 & 0.05 & 2 & 0.24 \\
11th & 2 & 0.05 & 2 & 0.24 \\
12th & 2 & 0.05 & 2 & 0.24 \\
13th & 2 & 0.05 & 2 & 0.24 \\
\midrule
\textbf{Total} & \textbf{4,036} & \textbf{100.00} & \textbf{820} & \textbf{100.00} \\
\midrule
Repeated & 1,931 & 47.84 & 416 & 50.73 \\
\newpage
\multicolumn{5}{@{}l}{\textbf{10-minute context}}\\*
Only & 1,264 & 31.32 & 256 & 31.22 \\
First & 985 & 24.41 & 169 & 20.61 \\
2nd & 903 & 22.37 & 160 & 19.51 \\
3rd & 428 & 10.60 & 80 & 9.76 \\
4th & 192 & 4.76 & 42 & 5.12 \\
5th & 104 & 2.58 & 33 & 4.02 \\
6th & 61 & 1.51 & 21 & 2.56 \\
7th & 40 & 0.99 & 18 & 2.20 \\
8th & 20 & 0.50 & 10 & 1.22 \\
9th & 11 & 0.27 & 7 & 0.85 \\
10th & 5 & 0.12 & 3 & 0.37 \\
11th & 4 & 0.10 & 3 & 0.37 \\
12th & 3 & 0.07 & 2 & 0.24 \\
13th & 1 & 0.02 & 1 & 0.12 \\
14th & 2 & 0.05 & 2 & 0.24 \\
15th & 1 & 0.02 & 1 & 0.12 \\
16th & 1 & 0.02 & 1 & 0.12 \\
17th & 2 & 0.05 & 2 & 0.24 \\
18th & 1 & 0.02 & 1 & 0.12 \\
19th & 2 & 0.05 & 2 & 0.24 \\
20th & 3 & 0.07 & 3 & 0.37 \\
21st & 2 & 0.05 & 2 & 0.24 \\
22nd & 1 & 0.02 & 1 & 0.12 \\
\midrule
\textbf{Total} & \textbf{4,036} & \textbf{100.00} & \textbf{820} & \textbf{100.00} \\
\midrule
Repeated & 2,772 & 68.68 & 564 & 68.78 \\
\end{longtable}
\noindent\textit{Notes.} Only means that the queried event type occurs once in the context. First means that it occurs multiple times and the target is the first occurrence. All other ranks are exact, with zero cells shown explicitly. Each duration is expanded to the highest observed rank in either cohort. Repeated summarizes all non-Only rows and is not an additional category. Shares are normalized separately within each cohort and duration, using 4,036 and 820 as the respective denominators. Rounding can cause displayed shares to sum to slightly above or below 100\%.
\endgroup

\clearpage
\begingroup
\setlength{\tabcolsep}{4pt}
\renewcommand{\arraystretch}{1.05}
\setlength{\LTcapwidth}{\linewidth}
\newlength{\soccerordinalwidth}
\setlength{\soccerordinalwidth}{\dimexpr\linewidth-8\tabcolsep\relax}
\begin{longtable}{@{}>{\raggedright\arraybackslash}p{.30\soccerordinalwidth}*{4}{>{\raggedleft\arraybackslash}p{.175\soccerordinalwidth}}@{}}
\caption{Exact occurrence-order distributions of the soccer source population and the selected soccer targets. This component-level comparison documents the distribution produced by the sampling procedure.}\label{tab:soccer-sampling-ordinal-composition}\\
\toprule
& \multicolumn{2}{c}{Full soccer} & \multicolumn{2}{c}{Selected soccer} \\
\cmidrule(lr){2-3}\cmidrule(l){4-5}
Target order & Count & Share (\%) & Count & Share (\%) \\
\midrule
\endfirsthead
\multicolumn{5}{@{}l}{\tablename\ \thetable\ (continued)}\\
\toprule
& \multicolumn{2}{c}{Full soccer} & \multicolumn{2}{c}{Selected soccer} \\
\cmidrule(lr){2-3}\cmidrule(l){4-5}
Target order & Count & Share (\%) & Count & Share (\%) \\
\midrule
\endhead
\midrule
\multicolumn{5}{r@{}}{Continued on next page}\\
\endfoot
\bottomrule
\endlastfoot
\multicolumn{5}{@{}l}{\textbf{1-minute context}}\\*
Only & 3,035 & 86.32 & 260 & 86.67 \\
First & 237 & 6.74 & 15 & 5.00 \\
2nd & 228 & 6.48 & 23 & 7.67 \\
3rd & 16 & 0.46 & 2 & 0.67 \\
\midrule
\textbf{Total} & \textbf{3,516} & \textbf{100.00} & \textbf{300} & \textbf{100.00} \\
\midrule
Repeated & 481 & 13.68 & 40 & 13.33 \\
\midrule
\multicolumn{5}{@{}l}{\textbf{3-minute context}}\\*
Only & 2,358 & 67.06 & 204 & 68.00 \\
First & 519 & 14.76 & 45 & 15.00 \\
2nd & 502 & 14.28 & 41 & 13.67 \\
3rd & 104 & 2.96 & 8 & 2.67 \\
4th & 27 & 0.77 & 2 & 0.67 \\
5th & 4 & 0.11 & 0 & 0.00 \\
6th & 1 & 0.03 & 0 & 0.00 \\
7th & 1 & 0.03 & 0 & 0.00 \\
\midrule
\textbf{Total} & \textbf{3,516} & \textbf{100.00} & \textbf{300} & \textbf{100.00} \\
\midrule
Repeated & 1,158 & 32.94 & 96 & 32.00 \\
\newpage
\multicolumn{5}{@{}l}{\textbf{5-minute context}}\\*
Only & 1,862 & 52.96 & 161 & 53.67 \\
First & 663 & 18.86 & 56 & 18.67 \\
2nd & 700 & 19.91 & 60 & 20.00 \\
3rd & 192 & 5.46 & 14 & 4.67 \\
4th & 63 & 1.79 & 4 & 1.33 \\
5th & 22 & 0.63 & 4 & 1.33 \\
6th & 10 & 0.28 & 0 & 0.00 \\
7th & 4 & 0.11 & 1 & 0.33 \\
\midrule
\textbf{Total} & \textbf{3,516} & \textbf{100.00} & \textbf{300} & \textbf{100.00} \\
\midrule
Repeated & 1,654 & 47.04 & 139 & 46.33 \\
\midrule
\multicolumn{5}{@{}l}{\textbf{10-minute context}}\\*
Only & 1,101 & 31.31 & 93 & 31.00 \\
First & 894 & 25.43 & 78 & 26.00 \\
2nd & 813 & 23.12 & 70 & 23.33 \\
3rd & 380 & 10.81 & 32 & 10.67 \\
4th & 162 & 4.61 & 12 & 4.00 \\
5th & 79 & 2.25 & 8 & 2.67 \\
6th & 44 & 1.25 & 4 & 1.33 \\
7th & 25 & 0.71 & 3 & 1.00 \\
8th & 10 & 0.28 & 0 & 0.00 \\
9th & 4 & 0.11 & 0 & 0.00 \\
10th & 2 & 0.06 & 0 & 0.00 \\
11th & 1 & 0.03 & 0 & 0.00 \\
12th & 1 & 0.03 & 0 & 0.00 \\
\midrule
\textbf{Total} & \textbf{3,516} & \textbf{100.00} & \textbf{300} & \textbf{100.00} \\
\midrule
Repeated & 2,415 & 68.69 & 207 & 69.00 \\
\end{longtable}
\noindent\textit{Notes.} Only means that the queried event type occurs once in the context. First means that it occurs multiple times and the target is the first occurrence. All other ranks are exact, with zero cells shown explicitly. Each duration is expanded to the highest observed rank in either cohort. Repeated summarizes all non-Only rows and is not an additional category. Shares are normalized separately within each cohort and duration, using 3,516 and 300 as the respective denominators. Rounding can cause displayed shares to sum to slightly above or below 100\%.
\endgroup

\newlength{\frozenevaltablewidth}

\clearpage
\subsection{Input Configurations of Evaluated Models}

\begingroup
\setlength{\tabcolsep}{2pt}
\renewcommand{\arraystretch}{1.12}
\setlength{\LTcapwidth}{\linewidth}
\setlength{\frozenevaltablewidth}{\dimexpr\linewidth-4\tabcolsep\relax}
\begin{longtable}{@{}>{\raggedright\arraybackslash}p{0.240\frozenevaltablewidth}>{\raggedright\arraybackslash}p{0.160\frozenevaltablewidth}>{\raggedright\arraybackslash}p{0.600\frozenevaltablewidth}@{}}
\caption{Input configurations used in the original Event-STVG evaluation.}\label{tab:appendix-model-inputs}\\
\toprule
Model & Settings & Input Configuration \\
\midrule
\endfirsthead
\multicolumn{3}{@{}l}{\tablename\ \thetable\ (continued)}\\
\toprule
Model & Settings & Input Configuration \\
\midrule
\endhead
\multicolumn{3}{r@{}}{Continued on next page}\\
\endfoot
\bottomrule
\endlastfoot
Gemini-3.7-Flash & Oracle / 1 / 3 / 5 / 10 min & 512\(\times\)288 JPEG, 2 FPS except for the 10-minute setting. Oracle: variable frames; 1/3/5 min: 2 FPS, 120/360/600 frames; 10 min: 900 uniformly sampled frames (\textasciitilde{}1.5 FPS). \\
GPT-5.6-Sol & Oracle / 1 / 3 / 5 / 10 min & 512\(\times\)288 JPEG, 2 FPS, \texttt{detail=low}. Oracle: variable frames; 1/3/5/10 min: 120/360/600/1200 frames. \\
Qwen3.8-Max & Oracle / 1 / 3 / 5 / 10 min & Native 1280\(\times\)720 MP4. \\
Qwen3.7-Plus & Oracle / 1 min & Native 1280\(\times\)720 MP4. \\
Qwen3.5-9B & Oracle / 1 min & Native 1280\(\times\)720 MP4. \\
GLM-5.3-Flash & Oracle / 1 min & Native 1280\(\times\)720 MP4. \\
GLM-5V-Turbo & Oracle / 1 min & Native 1280\(\times\)720 MP4. \\
GLM-4.6V & Oracle / 1 min & Native 1280\(\times\)720 MP4. \\
MiniMax-M3 & Oracle / 1 min & Native 1280\(\times\)720 MP4. \\
PTD (Qwen3-VL-4B) & Oracle / 1 min & Native 1280\(\times\)720 MP4. Uniform frame sampling; up to 64 frames for Oracle (all available frames if fewer than 64), and 64 frames for 1 min (\textasciitilde{}1 FPS). Preprocessed at 1152\(\times\)640. Each semantic role is processed independently. \\
STVG-R1-7B & Oracle / 1 min & Track-ID-annotated video; 1 FPS, up to 64 frames; total pixel budget of 6,422,528 with dynamic resizing. \\
VideoMind-7B + Spatial Backend & Oracle / 1 min & Native 1280\(\times\)720 MP4; Planner/Grounder: 1 FPS, up to 100/150 frames; Verifier: 2 FPS, up to 64 frames. Per-frame pixel budget: 28,224--50,176, with dynamic resizing. \\
VideoTemp-o3 + Spatial Backend & Oracle / 1 min & Native 1280\(\times\)720 MP4; initial sampling at 2 FPS, up to 1024 frames; per-frame pixel budget of 50,176 with dynamic resizing. Optional tool-based cropping at 2 FPS, nominally up to 64 frames. \\
\end{longtable}
\endgroup

\textit{Notes.} For general VLMs, we use native video input when supported by the model provider; otherwise, we use frame-based input. For Gemini-3.7-Flash and GPT-5.6-Sol, we adopt 2 FPS video sampling, consistent with the documented default sampling rate for Qwen video input. Frames are resized to 512\(\times\)288 to accommodate long-context input constraints. As an exception, Gemini uses 900 uniformly sampled frames (\textasciitilde{}1.5 FPS) at 10 minutes to stay within practical input limits and ensure successful inference.

The last three baselines use detection and tracking at 2 FPS with a maximum frame width of 960 pixels. Both agent-based methods share a Qwen2-VL role-selection backend (2 FPS, up to 64 frames, 28,224--100,352 pixels per frame). Oracle uses the provided target interval, while 1-minute spatial grounding uses model-predicted intervals. Actual frame counts depend on video duration and model-specific preprocessing. STVG-R1 uses four times its official default total pixel budget. VideoTemp-o3 did not invoke optional tools during Event-STVG evaluation.

\clearpage
\subsection{Evaluator and Scoring Rules}
\label{app:frozen-evaluator}
\paragraph{Frozen evaluator}
We use the fixed evaluator \texttt{role\_stvg\_eval\_v1.0} and response extractor \texttt{role\_stvg\_json\_extraction\_v1}. For each expected query, the last record in the accepted prediction file is authoritative, and its raw response is parsed again. Extraction first accepts a complete answer-shaped JSON value, then a whole-response JSON fence, and otherwise a unique embedded answer-shaped JSON occurrence. No output repair is applied. Surrounding prose and extra fields incur no additional formatting penalty when extraction succeeds. Missing predictions, API errors, and extraction failures remain in the evaluation population.

\paragraph{Response validation}
Context responses must contain both \texttt{temporal\_window} and \texttt{role\_localization}, whereas Oracle responses must be nonempty arrays of objects containing \texttt{time\_sec}. Successful extraction is followed by component-wise validation. After successful extraction, an empty or invalid localization container does not invalidate an otherwise valid temporal window: temporal IoU is retained, while joint and intersection-only grounding scores are zero. Invalid boxes or missing role keys affect only the corresponding role at that timestamp; duplicated required timestamps invalidate all roles at that timestamp. Explicit \texttt{null}, missing predictions, and invalid predictions are treated as distinct states.

\paragraph{Validation and temporal support}
A context prediction must explicitly provide a finite temporal window $[p_0,p_1]$ satisfying $0\leq p_0\leq p_1\leq D$, where $D$ is the input-video duration. An invalid window receives zero temporal and joint grounding scores; localization timestamps do not supply a replacement interval. Bounding boxes must contain four finite numeric coordinates in $[0,1000]$, with $x_{\min}<x_{\max}$ and $y_{\min}<y_{\max}$. Coordinates are interpreted directly as normalized \texttt{xyxy}; they are not clipped, reordered, or rescaled.

Oracle predictions are evaluated at the exact timestamps listed in the frozen prompt and ground truth. For context predictions, the required timestamp sequence is anchored at the predicted start, $p_0,p_0+0.5,\ldots\leq p_1$, with tolerance $10^{-6}$ seconds. Ground-truth timestamps are read from the annotated sequence rather than regenerated from its interval. Each row must match its own required sequence within this tolerance; extra timestamps are ignored, and missing timestamps remain missing. Duplicated rows at a required timestamp invalidate all roles at that timestamp. The validated sequences are aligned through the nominal 2-Hz index $q(t)=\lfloor 2t+0.5\rfloor$, with half ties rounded upward. No interpolation is used.

\paragraph{Temporal and joint grounding metrics}
Temporal IoU is computed on continuous intervals. For query $i$ with predicted interval $P_i$ and ground-truth interval $G_i$,
\begin{equation}
\operatorname{tIoU}_i=\frac{|P_i\cap G_i|}{|P_i\cup G_i|},
\end{equation}
with score zero for zero-length union. For queried role $r$, let $S_G$ and $S_P$ be the nominal ground-truth and predicted temporal supports, and let $Z_r$ contain their shared indices at which both role states are explicit \texttt{null}. The nullable extension of classic tube IoU is
\begin{equation}
v_{i,r}=\frac{\displaystyle\sum_{k\in S_G\cap S_P}
\mathbf{1}[g_{k,r},p_{k,r}\text{ are valid boxes}]
\operatorname{IoU}(g_{k,r},p_{k,r})}
{|S_G\cup S_P|-|Z_r|}.
\end{equation}
When the adjusted denominator is zero, the role score is one. A box versus \texttt{null}, a missing prediction, or an invalid prediction contributes zero numerator and retains the relevant denominator term. A shared explicit \texttt{null}/\texttt{null} pair is excluded rather than counted as a successful localized box. Predicted support outside the ground-truth support remains in the denominator even when the prediction is \texttt{null}. Oracle grounding uses the same rule with both supports fixed to the annotated timestamps.

\paragraph{Intersection-only diagnostic}
For \texttt{vIoU-Int}, the evaluation domain is $J=S_G\cap S_P$. For each role, let $G_r$ and $P_r$ be the indices in $J$ with valid ground-truth and predicted boxes, respectively, and let $E_r$ contain missing or invalid predictions. If the role has visible ground-truth support anywhere in the complete event, its intersection score is
\begin{equation}
v^{\mathrm{Int}}_{i,r}=\frac{\displaystyle\sum_{k\in G_r\cap P_r}
\operatorname{IoU}(g_{k,r},p_{k,r})}{|G_r\cup P_r\cup E_r|},
\end{equation}
with zero for an empty denominator. If the role is invisible throughout the complete event, its score is one only when $J$ is nonempty and every prediction in $J$ is an explicit \texttt{null}; otherwise it is zero. An empty intersection scores zero. Joint vIoU and vIoU-Int are computed independently; neither is obtained by multiplying the other by continuous temporal IoU.

\paragraph{Aggregation and failures}
Queried roles are matched by their semantic keys, without role permutation or best matching. Role scores are averaged equally within each query, and query scores are averaged equally over the fixed evaluation population:
\begin{equation}
v_i=\frac{1}{|R_i|}\sum_{r\in R_i}v_{i,r},\qquad
\operatorname{m\text{-}vIoU}=\frac{1}{N}\sum_{i=1}^{N}v_i.
\end{equation}
Mean temporal IoU and mean intersection IoU use the same equal-query convention. Failed queries retain zero scores in these means. Threshold rates are computed from the query-level mean role score and use strict comparison:
\begin{equation}
\operatorname{vIoU@}\theta=\frac{1}{N}\sum_{i=1}^{N}\mathbf{1}[v_i>\theta],
\qquad\theta\in\{0.3,0.5\}.
\end{equation}
Validation preserves component-level failures: a malformed role box affects that role and timestamp, whereas a valid temporal window can retain its tIoU even when spatial output is unavailable. Empty or invalid context localization containers receive zero spatial and joint grounding scores. Extra roles and fields do not create additional scored roles or additional penalties. All displayed percentages are the corresponding unit-interval scores multiplied by 100.

\clearpage
\subsection{Feasibility of Long-Context Video Inputs}

\textbf{Objective} We examine the input constraints of GPT-5.6-Sol and Gemini-3.7-Flash to explain the frame resolution and image detail settings adopted in our long-context evaluation.

\textbf{Setup} We randomly sample 30 events without replacement from the full 4,036-event dataset (seed = 21). Under the 10-minute setting, we test 1.5 FPS (900 frames) and 2 FPS (1,200 frames) at different resolutions. For GPT-5.6-Sol, the \texttt{detail} parameter is left unspecified.

\textbf{Results}

\begingroup
\setlength{\tabcolsep}{2pt}
\renewcommand{\arraystretch}{1.12}
\setlength{\LTcapwidth}{\linewidth}
\setlength{\frozenevaltablewidth}{\dimexpr\linewidth-10\tabcolsep\relax}
\begin{longtable}{@{}>{\raggedright\arraybackslash}p{0.200\frozenevaltablewidth}>{\raggedright\arraybackslash}p{0.150\frozenevaltablewidth}>{\raggedright\arraybackslash}p{0.160\frozenevaltablewidth}>{\raggedright\arraybackslash}p{0.100\frozenevaltablewidth}>{\raggedright\arraybackslash}p{0.160\frozenevaltablewidth}>{\raggedright\arraybackslash}p{0.230\frozenevaltablewidth}@{}}
\caption{Input feasibility under different resolutions and frame counts in the 10-minute setting.}\label{tab:appendix-input-feasibility}\\
\toprule
Model & Resolution & Detail & Frames & Successful responses & Main failure \\
\midrule
\endfirsthead
\multicolumn{6}{@{}l}{\tablename\ \thetable\ (continued)}\\
\toprule
Model & Resolution & Detail & Frames & Successful responses & Main failure \\
\midrule
\endhead
\multicolumn{6}{r@{}}{Continued on next page}\\
\endfoot
\bottomrule
\endlastfoot
GPT-5.6-Sol & 512\(\times\)288 & Unspecified & 1,200 & 0/30 & Primarily context limit \\
GPT-5.6-Sol & 1280\(\times\)720 & Unspecified & 1,200 & 0/30 & Primarily context limit \\
GPT-5.6-Sol & 512\(\times\)288 & Unspecified & 900 & 0/30 & Context limit \\
GPT-5.6-Sol & 1280\(\times\)720 & Unspecified & 900 & 0/30 & Context limit \\
Gemini-3.7-Flash & 512\(\times\)288 & --- & 1,200 & 0/30 & Request size / gateway limit \\
Gemini-3.7-Flash & 1280\(\times\)720 & --- & 1,200 & 0/30 & Request size / gateway limit \\
Gemini-3.7-Flash & 1280\(\times\)720 & --- & 900 & 0/30 & Request size / gateway limit \\
\end{longtable}
\endgroup

\textbf{Analysis} None of the tested configurations returned successful responses. GPT-5.6-Sol encountered context-limit errors even at 512\(\times\)288 when \texttt{detail} was unspecified, while Gemini-3.7-Flash was primarily constrained by request-size or gateway limits. These results motivate the use of resized frames and explicit image-detail settings to accommodate high frame counts in long-context evaluation.

\clearpage
\subsection{Effect of Frame Resolution on Grounding Performance}

\textbf{Objective} We examine whether increasing frame resolution improves Gemini-3.7-Flash's temporal and spatial grounding performance in the 1-minute Event-STVG setting.

\textbf{Setup} We select 100 events from the full 4,036-event dataset (seed = 21), stratified by target bounding-box area, width, and height using 20 quantile bins. The sample preserves the overall mean box statistics within 2\% and approximately matches the event occurrence distribution. It covers 37 of 45 event categories across all eight domains (80 soccer and 20 other-domain events). We compare 512\(\times\)288 and 1280\(\times\)720 frame inputs on the same events. We report paired differences (1280\(\times\)720 minus 512\(\times\)288, in percentage points) with 95\% confidence intervals estimated using source-clustered bootstrap across 55 video sources (20,000 resamples, seed = 21).

\textbf{Results}

\begingroup
\setlength{\tabcolsep}{2pt}
\renewcommand{\arraystretch}{1.12}
\setlength{\LTcapwidth}{\linewidth}
\setlength{\frozenevaltablewidth}{\dimexpr\linewidth-8\tabcolsep\relax}
\begin{longtable}{@{}>{\raggedright\arraybackslash}p{0.180\frozenevaltablewidth}>{\raggedright\arraybackslash}p{0.170\frozenevaltablewidth}>{\raggedright\arraybackslash}p{0.170\frozenevaltablewidth}>{\raggedright\arraybackslash}p{0.160\frozenevaltablewidth}>{\raggedright\arraybackslash}p{0.320\frozenevaltablewidth}@{}}
\caption{Gemini-3.7-Flash performance (\%) under different frame resolutions.}\label{tab:appendix-frame-resolution}\\
\toprule
Metric & 512\(\times\)288 & 1280\(\times\)720 & \(\Delta\) (pp) & 95\% CI \\
\midrule
\endfirsthead
\multicolumn{5}{@{}l}{\tablename\ \thetable\ (continued)}\\
\toprule
Metric & 512\(\times\)288 & 1280\(\times\)720 & \(\Delta\) (pp) & 95\% CI \\
\midrule
\endhead
\multicolumn{5}{r@{}}{Continued on next page}\\
\endfoot
\bottomrule
\endlastfoot
m-tIoU & 45.04 & 50.46 & +5.42 & [\(-\)0.09, +10.88] \\
m-vIoU & 13.24 & 16.48 & +3.24 & [\(-\)1.05, +7.54] \\
\end{longtable}
\endgroup

\textbf{Analysis} Higher resolution yields improvements of 5.42 pp in m-tIoU and 3.24 pp in m-vIoU, although both 95\% CIs include zero. Thus, we do not observe conclusive improvements from increasing frame resolution on this sample. Joint grounding performance remains low even at 1280\(\times\)720, suggesting that the difficulty of Event-STVG cannot be attributed solely to input resolution.

\clearpage
\subsection{Effect of Image Detail Setting}

\textbf{Objective} We examine whether explicitly setting \texttt{detail=low} affects GPT-5.6-Sol`s grounding performance compared with omitting the detail parameter.

\textbf{Setup} We conduct a paired evaluation on the 300 soccer events used for long-context analysis, using their 1-minute videos. Both configurations use 2 FPS, 512\(\times\)288 resized JPEG frames, with 120 frames per video and default generation parameters. The only change is whether \texttt{detail=low} is explicitly specified or omitted. We report paired differences (omitted minus low, in percentage points) with 95\% confidence intervals estimated using source-clustered bootstrap across 50 source matches (20,000 resamples, seed = 21; percentile intervals).

\textbf{Results}

\begingroup
\setlength{\tabcolsep}{2pt}
\renewcommand{\arraystretch}{1.12}
\setlength{\LTcapwidth}{\linewidth}
\setlength{\frozenevaltablewidth}{\dimexpr\linewidth-8\tabcolsep\relax}
\begin{longtable}{@{}>{\raggedright\arraybackslash}p{0.180\frozenevaltablewidth}>{\raggedright\arraybackslash}p{0.170\frozenevaltablewidth}>{\raggedright\arraybackslash}p{0.170\frozenevaltablewidth}>{\raggedright\arraybackslash}p{0.160\frozenevaltablewidth}>{\raggedright\arraybackslash}p{0.320\frozenevaltablewidth}@{}}
\caption{Effect of the image detail setting on 1-minute grounding performance (\%).}\label{tab:appendix-gpt-detail}\\
\toprule
Metric & detail=low & Omitted & \(\Delta\) (pp) & 95\% CI \\
\midrule
\endfirsthead
\multicolumn{5}{@{}l}{\tablename\ \thetable\ (continued)}\\
\toprule
Metric & detail=low & Omitted & \(\Delta\) (pp) & 95\% CI \\
\midrule
\endhead
\multicolumn{5}{r@{}}{Continued on next page}\\
\endfoot
\bottomrule
\endlastfoot
m-vIoU & 2.21 & 2.02 & \(-\)0.19 & [\(-\)0.60, +0.25] \\
m-tIoU & 45.14 & 45.32 & +0.18 & [\(-\)1.55, +2.00] \\
\end{longtable}
\endgroup

\textbf{Analysis} Omitting the detail parameter does not yield a conclusive improvement in any of the two metrics, with all 95\% CIs including zero. Both configurations achieve similar temporal grounding scores (around 45\% m-tIoU), while joint grounding remains low (around 2\% m-vIoU). A QA-level bootstrap sensitivity analysis leads to the same conclusion. These results suggest that explicitly setting \texttt{detail=low} does not explain the low joint grounding performance observed under this 1-minute evaluation setting.

\clearpage
\subsection{Input Format Alignment in Short-Context Evaluation}

\textbf{Objective} To improve the comparability of general-purpose VLMs, we re-evaluate seven models originally tested with native video input using a unified frame-based input protocol and examine the resulting performance differences.

\textbf{Setup} We sample 100 events from the full 4,036-event dataset (seed = 21), preserving the event distribution (87 soccer and 13 other-domain events). The sample covers 53 video sources. For frame-based input, we follow the Gemini input protocol, sampling at 2 FPS and resizing frames to 512 \(\times\) 288. Both input formats are evaluated on the same events under Oracle and 1-minute settings. We report paired differences (frames minus native, in percentage points) with 95\% confidence intervals estimated using source-clustered bootstrap (20,000 resamples, seed = 21).

\textbf{Results}

\begingroup
\setlength{\tabcolsep}{2pt}
\renewcommand{\arraystretch}{1.12}
\setlength{\LTcapwidth}{\linewidth}
\setlength{\LTpre}{3pt}
\setlength{\LTpost}{3pt}
\setlength{\frozenevaltablewidth}{\dimexpr\linewidth-8\tabcolsep\relax}
\begin{longtable}{@{}>{\raggedright\arraybackslash}p{0.240\frozenevaltablewidth}>{\raggedright\arraybackslash}p{0.130\frozenevaltablewidth}>{\raggedright\arraybackslash}p{0.130\frozenevaltablewidth}>{\raggedright\arraybackslash}p{0.140\frozenevaltablewidth}>{\raggedright\arraybackslash}p{0.360\frozenevaltablewidth}@{}}
\caption{Oracle m-vIoU (\%) under native video and frame-based inputs.}\label{tab:appendix-short-alignment-oracle}\\
\toprule
Model & Native & Frames & \(\Delta\) (pp) & 95\% CI \\
\midrule
\endfirsthead
\multicolumn{5}{@{}l}{\tablename\ \thetable\ (continued)}\\
\toprule
Model & Native & Frames & \(\Delta\) (pp) & 95\% CI \\
\midrule
\endhead
\multicolumn{5}{r@{}}{Continued on next page}\\
\endfoot
\bottomrule
\endlastfoot
Qwen3.8-Max & 12.87 & 27.81 & +14.94 & [+10.58, +19.59] \\
MiniMax-M3 & 1.08 & 8.51 & +7.43 & [+5.01, +10.19] \\
Qwen3.7-Plus & 3.60 & 21.83 & +18.23 & [+14.72, +21.76] \\
GLM-4.6V & 12.73 & 10.68 & \(-\)2.05 & [\(-\)4.84, +0.76] \\
GLM-5V-Turbo & 9.60 & 14.19 & +4.60 & [+1.83, +7.41] \\
GLM-5.3-Flash & 11.40 & 12.13 & +0.73 & [\(-\)2.23, +3.77] \\
Qwen3.5-9B & 3.36 & 3.59 & +0.23 & [\(-\)1.42, +2.07] \\
\end{longtable}
\endgroup

\begingroup
\setlength{\tabcolsep}{2pt}
\renewcommand{\arraystretch}{1.12}
\setlength{\LTcapwidth}{\linewidth}
\setlength{\LTpre}{3pt}
\setlength{\LTpost}{3pt}
\setlength{\frozenevaltablewidth}{\dimexpr\linewidth-8\tabcolsep\relax}
\begin{longtable}{@{}>{\raggedright\arraybackslash}p{0.240\frozenevaltablewidth}>{\raggedright\arraybackslash}p{0.130\frozenevaltablewidth}>{\raggedright\arraybackslash}p{0.130\frozenevaltablewidth}>{\raggedright\arraybackslash}p{0.140\frozenevaltablewidth}>{\raggedright\arraybackslash}p{0.360\frozenevaltablewidth}@{}}
\caption{1-minute m-vIoU (\%) under native video and frame-based inputs.}\label{tab:appendix-short-alignment-viou}\\
\toprule
Model & Native & Frames & \(\Delta\) (pp) & 95\% CI \\
\midrule
\endfirsthead
\multicolumn{5}{@{}l}{\tablename\ \thetable\ (continued)}\\
\toprule
Model & Native & Frames & \(\Delta\) (pp) & 95\% CI \\
\midrule
\endhead
\multicolumn{5}{r@{}}{Continued on next page}\\
\endfoot
\bottomrule
\endlastfoot
Qwen3.8-Max & 3.36 & 6.43 & +3.07 & [+0.87, +5.66] \\
MiniMax-M3 & 0.02 & 1.19 & +1.17 & [+0.47, +2.22] \\
Qwen3.7-Plus & 1.07 & 1.99 & +0.92 & [+0.24, +1.78] \\
GLM-4.6V & 0.83 & 0.74 & \(-\)0.09 & [\(-\)1.60, +1.01] \\
GLM-5V-Turbo & 1.40 & 1.26 & \(-\)0.14 & [\(-\)1.24, +0.70] \\
GLM-5.3-Flash & 1.31 & 1.45 & +0.14 & [\(-\)0.85, +1.13] \\
Qwen3.5-9B & 0.11 & 0.34 & +0.23 & [\(-\)0.11, +0.90] \\
\end{longtable}
\endgroup

\begingroup
\setlength{\tabcolsep}{2pt}
\renewcommand{\arraystretch}{1.12}
\setlength{\LTcapwidth}{\linewidth}
\setlength{\LTpre}{3pt}
\setlength{\LTpost}{3pt}
\setlength{\frozenevaltablewidth}{\dimexpr\linewidth-8\tabcolsep\relax}
\begin{longtable}{@{}>{\raggedright\arraybackslash}p{0.240\frozenevaltablewidth}>{\raggedright\arraybackslash}p{0.130\frozenevaltablewidth}>{\raggedright\arraybackslash}p{0.130\frozenevaltablewidth}>{\raggedright\arraybackslash}p{0.140\frozenevaltablewidth}>{\raggedright\arraybackslash}p{0.360\frozenevaltablewidth}@{}}
\caption{1-minute m-tIoU (\%) under native video and frame-based inputs.}\label{tab:appendix-short-alignment-tiou}\\
\toprule
Model & Native & Frames & \(\Delta\) (pp) & 95\% CI \\
\midrule
\endfirsthead
\multicolumn{5}{@{}l}{\tablename\ \thetable\ (continued)}\\
\toprule
Model & Native & Frames & \(\Delta\) (pp) & 95\% CI \\
\midrule
\endhead
\multicolumn{5}{r@{}}{Continued on next page}\\
\endfoot
\bottomrule
\endlastfoot
Qwen3.8-Max & 45.12 & 41.17 & \(-\)3.95 & [\(-\)11.13, +2.94] \\
MiniMax-M3 & 4.17 & 26.36 & +22.19 & [+16.14, +28.47] \\
Qwen3.7-Plus & 34.02 & 31.75 & \(-\)2.26 & [\(-\)7.15, +2.52] \\
GLM-4.6V & 9.17 & 12.78 & +3.60 & [\(-\)2.43, +9.99] \\
GLM-5V-Turbo & 16.75 & 15.84 & \(-\)0.92 & [\(-\)6.50, +4.30] \\
GLM-5.3-Flash & 33.39 & 31.71 & \(-\)1.68 & [\(-\)9.46, +6.04] \\
Qwen3.5-9B & 15.19 & 18.47 & +3.27 & [\(-\)3.18, +9.80] \\
\end{longtable}
\endgroup

\textbf{Analysis} Frame-based input improves Oracle m-vIoU for four of the seven models, with 95\% CIs excluding zero. At 1 minute, such improvements are observed for only three models (Qwen3.8-Max, MiniMax-M3, and Qwen3.7-Plus), with smaller gains. For temporal grounding, only MiniMax-M3 shows a clear improvement (+22.19 pp), while the other models exhibit no conclusive changes. Notably, all six general-purpose VLMs excluded from our long-context evaluation in Sec. 4.2 still achieve below 2\% m-vIoU after input alignment. These results further support our model selection for long-context evaluation, while showing that input alignment does not fundamentally change the observed difficulty of joint spatio-temporal grounding.

\clearpage
\subsection{Input Format Alignment in Long-Context Evaluation}

\textbf{Objective} To improve the comparability of general-purpose VLMs in long-context evaluation, we re-evaluate Qwen3.8-Max, originally tested with native video input, using the exact frame-based input protocol adopted for Gemini.

\textbf{Setup} We sample 100 events from the full 820-event long-context evaluation set (seed = 21), preserving the event distribution (36 soccer and 64 other-domain events). The sample covers 64 video sources. Both input formats are evaluated on the same events under 3-, 5-, and 10-minute settings. We report paired differences (frames minus native, in percentage points) with 95\% confidence intervals estimated using source-clustered bootstrap (20,000 resamples, seed = 21).

\textbf{Results}

\begingroup
\setlength{\tabcolsep}{2pt}
\renewcommand{\arraystretch}{1.12}
\setlength{\LTcapwidth}{\linewidth}
\setlength{\frozenevaltablewidth}{\dimexpr\linewidth-10\tabcolsep\relax}
\begin{longtable}{@{}>{\raggedright\arraybackslash}p{0.130\frozenevaltablewidth}>{\raggedright\arraybackslash}p{0.150\frozenevaltablewidth}>{\raggedright\arraybackslash}p{0.130\frozenevaltablewidth}>{\raggedright\arraybackslash}p{0.130\frozenevaltablewidth}>{\raggedright\arraybackslash}p{0.140\frozenevaltablewidth}>{\raggedright\arraybackslash}p{0.320\frozenevaltablewidth}@{}}
\caption{Qwen3.8-Max performance (\%) under native video and frame-based inputs across long-context settings.}\label{tab:appendix-long-alignment}\\
\toprule
Duration & Metric & Native & Frames & \(\Delta\) (pp) & 95\% CI \\
\midrule
\endfirsthead
\multicolumn{6}{@{}l}{\tablename\ \thetable\ (continued)}\\
\toprule
Duration & Metric & Native & Frames & \(\Delta\) (pp) & 95\% CI \\
\midrule
\endhead
\multicolumn{6}{r@{}}{Continued on next page}\\
\endfoot
\bottomrule
\endlastfoot
3 min & m-vIoU & 2.88 & 4.37 & +1.49 & [\(-\)0.61, +3.69] \\
3 min & m-tIoU & 38.18 & 35.03 & \(-\)3.15 & [\(-\)9.46, +3.16] \\
5 min & m-vIoU & 3.11 & 3.72 & +0.61 & [\(-\)2.18, +3.28] \\
5 min & m-tIoU & 31.27 & 32.01 & +0.74 & [\(-\)5.81, +6.96] \\
10 min & m-vIoU & 1.91 & 2.16 & +0.25 & [\(-\)2.00, +2.76] \\
10 min & m-tIoU & 26.48 & 29.82 & +3.34 & [\(-\)3.86, +10.74] \\
\end{longtable}
\endgroup

\textbf{Analysis} Frame-based input yields slightly higher m-vIoU across all three durations, but all 95\% CIs include zero. Similarly, no conclusive differences are observed in m-tIoU. These results are consistent with our main findings that joint spatio-temporal grounding remains challenging in long-context Event-STVG evaluation.

\textbf{Event-Instance Confusion} A wrong-instance selection occurs when the prediction misses the target interval but overlaps another valid occurrence of the same event type. The error rates are 5.26\% vs. 7.89\% at 3 minutes, 5.26\% vs. 2.63\% at 5 minutes, and 18.42\% vs. 13.16\% at 10 minutes (native vs. frames). No consistent improvement is observed across durations, and the 10-minute difference is inconclusive (\(-\)5.26 pp, 95\% CI [\(-\)18.75, +5.71]).

\textbf{Complete Multi-Role Grounding} We further examine 40 multi-role queries selected from the same 100 events based solely on ground-truth annotations. Complete success requires every queried role to achieve trajectory vIoU \(>\) 0.3. Under Oracle, success increases from 5.0\% to 15.0\% (95\% CI for the difference: [+2.38, +20.45] pp). However, at 1 minute, success is only 0\% and 2.5\% (native and frames), and neither input achieves any complete success at 3, 5, or 10 minutes. Even among queries where both inputs achieve tIoU \(>\) 0.5 (11, 10, and 7 queries at 3, 5, and 10 minutes, respectively), complete multi-role success remains zero.

Overall, input alignment improves some aggregate scores and complete multi-role grounding under Oracle, but does not eliminate event-instance confusion or the difficulty of complete multi-role grounding in long-context Event-STVG.

\clearpage
\subsection{Human Baseline of the 10-Minute Setting}
\label{app:human-ten-minute}

\textbf{Objective} To complement the assisted cross-annotation verification, we evaluate human performance on Event-STVG through end-to-end task execution, providing an additional assessment of task feasibility and a reference for interpreting model performance.

\textbf{Setup} We select 30 queries from the 820-query long-context evaluation pool (300 soccer and 520 cross-domain queries) using constrained randomized sampling (seed = 21). The occurrence distribution is 5 Only, 5 First, 5 second, 4 third, 4 fourth--fifth, 4 sixth--tenth, and 3 eleventh-or-later occurrences. The subset also includes 19 single-role, 6 two-role, and 5 three-role queries. Sampling is independent of model predictions and performance.

Six annotators independent of the research team participate in the evaluation, with three assigned to each resolution setting (original and 512\(\times\)288). All annotators evaluate the same 30 ten-minute queries without access to ground-truth event intervals or candidate tracks. We report event hit rate, m-tIoU, and m-vIoU.

\textbf{Results}

\begingroup
\setlength{\tabcolsep}{2pt}
\renewcommand{\arraystretch}{1.12}
\setlength{\LTcapwidth}{\linewidth}
\setlength{\frozenevaltablewidth}{\dimexpr\linewidth-8\tabcolsep\relax}
\begin{longtable}{@{}>{\raggedright\arraybackslash}p{0.25\frozenevaltablewidth}>{\raggedright\arraybackslash}p{0.12\frozenevaltablewidth}>{\raggedright\arraybackslash}p{0.27\frozenevaltablewidth}>{\raggedright\arraybackslash}p{0.18\frozenevaltablewidth}>{\raggedright\arraybackslash}p{0.18\frozenevaltablewidth}@{}}
\caption{Human performance (\%) under original and reduced video resolutions.}\label{tab:human-ten-minute}\\
\toprule
Resolution & Annotator & Hit rate & m-tIoU & m-vIoU \\
\midrule
\endfirsthead
\multicolumn{5}{@{}l}{\tablename\ \thetable\ (continued)}\\
\toprule
Resolution & Annotator & Hit rate & m-tIoU & m-vIoU \\
\midrule
\endhead
\multicolumn{5}{r@{}}{Continued on next page}\\
\endfoot
\bottomrule
\endlastfoot
Original & 1 & 28/30 (93.33\%) & 79.87 & 60.70 \\
Original & 2 & 29/30 (96.67\%) & 86.61 & 70.15 \\
Original & 3 & 29/30 (96.67\%) & 88.23 & 69.70 \\
\textbf{Original mean} &  & \textbf{95.56\%} & \textbf{84.90} & \textbf{66.85} \\
512\(\times\)288 & 4 & 29/30 (96.67\%) & 84.18 & 68.19 \\
512\(\times\)288 & 5 & 29/30 (96.67\%) & 83.20 & 65.73 \\
512\(\times\)288 & 6 & 30/30 (100\%) & 86.13 & 71.49 \\
\textbf{512\(\times\)288 mean} &  & \textbf{97.78\%} & \textbf{84.50} & \textbf{68.47} \\
\end{longtable}
\endgroup

\textit{Means are computed across the three annotators in each condition.}

\textbf{Analysis} Human annotators achieve high event hit rates at both original and 512\(\times\)288 resolutions (95.56\% and 97.78\%). Their mean tIoU scores are 84.90\% and 84.50\%, with tIoU conditioned on successful hits reaching 88.85\% and 86.42\%, respectively. These scores are also close to the 89.74\% temporal agreement observed in assisted cross-annotation, although the evaluation protocols differ. Mean vIoU reaches 66.85\% and 68.47\%, respectively, substantially above the model performance reported in our main experiments. Overall, human performance remains high under both resolutions, suggesting that Event-STVG is tractable for humans even with reduced-resolution input.

\clearpage
\subsection{Run-to-Run Stability of Grounding Performance}

\textbf{Objective} We examine the reproducibility of Event-STVG performance across repeated inference runs, considering both model output variation and service-side failures.

\textbf{Setup} We randomly sample 352 events without replacement from the 3,516-event soccer subset (seed = 21), covering 16 event categories and 50 matches. The same events are evaluated under Oracle and 1-minute settings for four API models (Qwen3.8-Max, Qwen3.7-Plus, GLM-4.6V, and MiniMax-M3) and PTD-64f. R0 denotes the original benchmark run, while R1 and R2 are two additional runs under identical inference configurations. The API models use provider-default generation settings, and PTD uses fixed 64-frame sampling with greedy decoding.

All runs use the same ground truth and evaluator, retaining failed or invalid predictions as zero scores. We report paired differences relative to R0 with 95\% confidence intervals estimated using match-clustered bootstrap (50 matches, 20,000 resamples, seed = 21). Each resample retains all selected events from each sampled match, with scores averaged equally across events. The intervals are pointwise and are not adjusted for multiple comparisons.

\textbf{Results}

\begingroup
\setlength{\tabcolsep}{2pt}
\renewcommand{\arraystretch}{1.12}
\setlength{\LTcapwidth}{\linewidth}
\setlength{\frozenevaltablewidth}{\dimexpr\linewidth-14\tabcolsep\relax}
\begin{longtable}{@{}>{\raggedright\arraybackslash}p{0.157\frozenevaltablewidth}>{\raggedright\arraybackslash}p{0.087\frozenevaltablewidth}>{\raggedright\arraybackslash}p{0.103\frozenevaltablewidth}>{\raggedright\arraybackslash}p{0.095\frozenevaltablewidth}>{\raggedright\arraybackslash}p{0.095\frozenevaltablewidth}>{\raggedright\arraybackslash}p{0.095\frozenevaltablewidth}>{\raggedright\arraybackslash}p{0.184\frozenevaltablewidth}>{\raggedright\arraybackslash}p{0.184\frozenevaltablewidth}@{}}
\caption{Grounding performance (\%) across three inference runs. R0 denotes the original run; R1 and R2 are additional runs. Differences and 95\% CIs are in percentage points.}\label{tab:appendix-repeatability}\\
\toprule
Model & Setting & Metric & R0 & R1 & R2 & R1\(-\)R0 [95\% CI] & R2\(-\)R0 [95\% CI] \\
\midrule
\endfirsthead
\multicolumn{8}{@{}l}{\tablename\ \thetable\ (continued)}\\
\toprule
Model & Setting & Metric & R0 & R1 & R2 & R1\(-\)R0 [95\% CI] & R2\(-\)R0 [95\% CI] \\
\midrule
\endhead
\multicolumn{8}{r@{}}{Continued on next page}\\
\endfoot
\bottomrule
\endlastfoot
Qwen3.8-Max & Oracle & m-vIoU & 11.548 & 12.208 & 11.389 & +0.660 [\(-\)1.257, +2.581] & \(-\)0.159 [\(-\)1.779, +1.541] \\
Qwen3.8-Max & 1 min & m-tIoU & 47.726 & 46.905 & 44.702 & \(-\)0.822 [\(-\)3.708, +2.051] & \(-\)3.024 [\(-\)5.686, \(-\)0.445] \\
Qwen3.8-Max & 1 min & m-vIoU & 3.384 & 3.057 & 3.756 & \(-\)0.327 [\(-\)1.026, +0.395] & +0.372 [\(-\)0.256, +1.028] \\
Qwen3.7-Plus & Oracle & m-vIoU & 2.518 & 2.654 & 2.235 & +0.136 [\(-\)0.386, +0.677] & \(-\)0.284 [\(-\)0.652, +0.108] \\
Qwen3.7-Plus & 1 min & m-tIoU & 38.095 & 38.346 & 41.125 & +0.250 [\(-\)2.073, +2.575] & +3.030 [+0.509, +5.700] \\
Qwen3.7-Plus & 1 min & m-vIoU & 0.383 & 0.418 & 0.616 & +0.035 [\(-\)0.108, +0.184] & +0.233 [+0.044, +0.442] \\
GLM-4.6V & Oracle & m-vIoU & 10.889 & 7.592 & 6.481 & \(-\)3.296 [\(-\)4.427, \(-\)2.185] & \(-\)4.408 [\(-\)5.686, \(-\)3.205] \\
GLM-4.6V & 1 min & m-tIoU & 7.550 & 4.658 & 4.190 & \(-\)2.893 [\(-\)5.296, \(-\)0.574] & \(-\)3.360 [\(-\)5.258, \(-\)1.433] \\
GLM-4.6V & 1 min & m-vIoU & 0.159 & 0.119 & 0.071 & \(-\)0.039 [\(-\)0.139, +0.035] & \(-\)0.088 [\(-\)0.211, +0.015] \\
MiniMax-M3 & Oracle & m-vIoU & 0.262 & 0.269 & 0.198 & +0.007 [\(-\)0.170, +0.149] & \(-\)0.065 [\(-\)0.213, +0.057] \\
MiniMax-M3 & 1 min & m-tIoU & 3.552 & 5.154 & 3.998 & +1.602 [\(-\)0.008, +3.243] & +0.445 [\(-\)1.027, +1.935] \\
MiniMax-M3 & 1 min & m-vIoU & 0.019 & 0.026 & 0.003 & +0.007 [\(-\)0.024, +0.046] & \(-\)0.016 [\(-\)0.037, +0.00005] \\
PTD-64f & Oracle & m-vIoU & 13.736 & 13.736 & 13.736 & 0.000 [0.000, 0.000] & 0.000 [0.000, 0.000] \\
PTD-64f & 1 min & m-tIoU & 15.719 & 15.719 & 15.719 & 0.000 [0.000, 0.000] & 0.000 [0.000, 0.000] \\
PTD-64f & 1 min & m-vIoU & 1.659 & 1.659 & 1.659 & 0.000 [0.000, 0.000] & 0.000 [0.000, 0.000] \\
\end{longtable}
\endgroup

\textbf{Analysis} The two Qwen models show relatively small variation in Oracle m-vIoU, with all 95\% CIs including zero. At 1 minute, Qwen3.8-Max's m-tIoU decreases by 3.02 pp from R0 to R2 (95\% CI [\(-\)5.69, \(-\)0.45]), while Qwen3.7-Plus's m-tIoU and m-vIoU increase by 3.03 pp ([+0.51, +5.70]) and 0.23 pp ([+0.04, +0.44]), respectively.

GLM-4.6V exhibits larger declines, accompanied by increased service failures. Its parseable Oracle predictions decrease from 352 in R0 to 217 and 200 in R1 and R2. On the 124 queries successfully parsed in all three runs, Oracle m-vIoU varies only from 12.52\% to 13.44\%, suggesting that service failures contribute substantially to the full-set differences. MiniMax-M3 maintains very low spatial grounding scores across runs, while PTD produces identical outputs under its fixed 64-frame greedy configuration.

\textbf{Despite these variations, the relative ranking of the five evaluated models by 1-minute m-vIoU remains unchanged across all three runs.} Qwen3.8-Max consistently outperforms the other three general-purpose VLMs, whose m-vIoU scores all remain below 1\%. These results further support our model selection for long-context evaluation in Sec. 4.2, while showing that the main observations are not altered by the measured run-to-run variations.

\clearpage
\subsection{Effect of PTD's Task Interface Adaptation}

\textbf{Objective} To evaluate PTD on Event-STVG, we adapt event queries to its original referring-query interface and retain its native output format to obtain valid grounding predictions. We examine the effect of this adaptation by first providing PTD with the full Event-STVG task instructions, and then testing the PTD-style interface on Qwen3.8-Max.

\textbf{PTD-style prompt} For each required semantic role, we construct a referring query by removing one terminal period from its frozen role definition and lowercasing only the first character. The query is inserted into PTD's native prompt template:

\begin{quote}
Given the query: '\{query\}' Localize the described object throughout the video. Use object reference tokens, time tokens, and box tokens. Return the object reference, event time segment, and per-time bbox coordinates.
\end{quote}

For a \textit{Shot on Target} event, the role-specific queries are:
\begin{enumerate}
\renewcommand{\labelenumi}{\arabic{enumi})}
\item \texttt{last\_passer}: the attacking player who makes the final effective pass to the shooter before the shot
\item \texttt{shooter}: the attacking player who takes the shot
\item \texttt{goalkeeper}: the goalkeeper defending the goal against the shot
\end{enumerate}

This construction uses the frozen role definitions without sample-specific rewriting or ground-truth-dependent query construction. The short queries do not explicitly retain the target occurrence order or the complete event-boundary specification. In the Qwen3.8-Max PTD-style condition, we use the same role-specific referring queries with simplified JSON outputs.

\textbf{Setup} We sample 100 events from the 3,516-event soccer subset using event-type-stratified sampling (seed = 21), covering 14 event categories and 46 matches. Both comparisons use Oracle and 1-minute settings.

For PTD, we compare its original short referring query with a full-information prompt that includes Event-STVG task instructions while retaining the native grounding output wrapper. Both conditions use independent role-wise inference (183 requests per setting), with identical model weights, decoding settings, 64-frame sampling configurations, and sampled source-frame indices.

\textbf{PTD implementation} PTD uses FP16 inference with SDPA attention, a single GPU worker, greedy decoding, and a maximum of 1,024 generated tokens. Each semantic role is processed independently using the same sampled source frames. Oracle inputs retain all distinct frames when at most 64 are available and otherwise use 64 uniformly sampled distinct frames. One-minute inputs use 64 uniformly sampled distinct frames spanning the complete video, including its first and last frames.

The PTD output adapter maps sampled-position time tokens back to video timestamps using the sampled frame indices. Sparse box predictions are interpolated only between valid neighboring predictions onto the 0.5-second evaluation grid, without temporal extrapolation. Failed role outputs are retained as missing predictions. This conversion is performed before scoring; the frozen evaluator itself does not interpolate trajectories.

For Qwen3.8-Max, we compare the original full prompt, which jointly requests all roles, against the PTD-style interface, which uses short referring queries, independent role-wise requests (183 per setting), and simplified JSON outputs. Native video input and provider-default inference settings remain unchanged.

All predictions use the same ground truth and evaluator, with unsuccessful outputs retained in the denominator. We report paired differences with 95\% confidence intervals estimated using match-clustered bootstrap (20,000 resamples across 46 matches, seed = 21).

\clearpage
\textbf{Results}

\begingroup
\setlength{\tabcolsep}{2pt}
\renewcommand{\arraystretch}{1.12}
\setlength{\LTcapwidth}{\linewidth}
\setlength{\frozenevaltablewidth}{\dimexpr\linewidth-10\tabcolsep\relax}
\begin{longtable}{@{}>{\raggedright\arraybackslash}p{0.140\frozenevaltablewidth}>{\raggedright\arraybackslash}p{0.150\frozenevaltablewidth}>{\raggedright\arraybackslash}p{0.170\frozenevaltablewidth}>{\raggedright\arraybackslash}p{0.180\frozenevaltablewidth}>{\raggedright\arraybackslash}p{0.130\frozenevaltablewidth}>{\raggedright\arraybackslash}p{0.230\frozenevaltablewidth}@{}}
\caption{PTD performance (\%) under the original referring-query and full-information interfaces. \(\Delta\) = Full information \(-\) Original query.}\label{tab:appendix-ptd-full-information}\\
\toprule
Setting & Metric & Original query & Full information & \(\Delta\) (pp) & 95\% CI \\
\midrule
\endfirsthead
\multicolumn{6}{@{}l}{\tablename\ \thetable\ (continued)}\\
\toprule
Setting & Metric & Original query & Full information & \(\Delta\) (pp) & 95\% CI \\
\midrule
\endhead
\multicolumn{6}{r@{}}{Continued on next page}\\
\endfoot
\bottomrule
\endlastfoot
Oracle & m-vIoU & 13.04 & 10.37 & \(-\)2.67 & [\(-\)5.08, \(-\)0.31] \\
1 min & m-tIoU & 16.16 & 9.72 & \(-\)6.45 & [\(-\)10.38, \(-\)2.36] \\
1 min & m-vIoU & 2.10 & 1.22 & \(-\)0.88 & [\(-\)1.78, \(-\)0.17] \\
\end{longtable}
\endgroup

\begingroup
\setlength{\tabcolsep}{2pt}
\renewcommand{\arraystretch}{1.12}
\setlength{\LTcapwidth}{\linewidth}
\setlength{\frozenevaltablewidth}{\dimexpr\linewidth-10\tabcolsep\relax}
\begin{longtable}{@{}>{\raggedright\arraybackslash}p{0.140\frozenevaltablewidth}>{\raggedright\arraybackslash}p{0.150\frozenevaltablewidth}>{\raggedright\arraybackslash}p{0.170\frozenevaltablewidth}>{\raggedright\arraybackslash}p{0.180\frozenevaltablewidth}>{\raggedright\arraybackslash}p{0.130\frozenevaltablewidth}>{\raggedright\arraybackslash}p{0.230\frozenevaltablewidth}@{}}
\caption{Qwen3.8-Max performance (\%) under the full-prompt and PTD-style interfaces. \(\Delta\) = PTD-style \(-\) Full prompt.}\label{tab:appendix-ptd-interface}\\
\toprule
Setting & Metric & Full prompt & PTD-style & \(\Delta\) (pp) & 95\% CI \\
\midrule
\endfirsthead
\multicolumn{6}{@{}l}{\tablename\ \thetable\ (continued)}\\
\toprule
Setting & Metric & Full prompt & PTD-style & \(\Delta\) (pp) & 95\% CI \\
\midrule
\endhead
\multicolumn{6}{r@{}}{Continued on next page}\\
\endfoot
\bottomrule
\endlastfoot
Oracle & m-vIoU & 9.45 & 8.53 & \(-\)0.92 & [\(-\)3.08, +1.33] \\
1 min & m-tIoU & 53.02 & 31.34 & \(-\)21.68 & [\(-\)27.24, \(-\)15.99] \\
1 min & m-vIoU & 4.21 & 2.18 & \(-\)2.03 & [\(-\)3.46, \(-\)0.63] \\
\end{longtable}
\endgroup

\textbf{Analysis} Providing full task instructions to PTD reduces all three reported scores, with 95\% CIs excluding zero. Notably, the number of native outputs marked OK also drops from 183 to 111 under Oracle and from 180 to 51 at 1 minute (out of 183 requests). For Qwen3.8-Max, adopting the PTD-style interface yields no conclusive difference under Oracle, but reduces both m-tIoU and m-vIoU at 1 minute.

Although the task-interface adaptations can affect performance comparisons, the substantial increase in output failures suggests that retaining PTD's original referring-query interface is a practical compromise to obtain valid predictions.

\clearpage
\subsection{Evaluation and Validation of STVG-R1-7B, VideoMind-7B, and VideoTemp-o3}

\textbf{Evaluation Protocol} We adapt STVG-R1-7B, VideoMind-7B, and VideoTemp-o3 to Event-STVG while retaining their original temporal grounding mechanisms. Unlike PTD, all models receive the complete Event-STVG query, including event definitions, occurrence order, and semantic roles.

\textbf{Baseline Adaptation} STVG-R1 operates on videos with automatically generated track IDs and predicts temporal boundaries and role identities. Its selected tracks are converted into bounding-box trajectories using SAM2 masks. VideoMind retains its Planner, Grounder, and Verifier modules, while VideoTemp retains its native agent loop with optional video-cropping tools. Since neither agent directly produces the required role trajectories, we use a shared spatial backend consisting of YOLOv12x, SAM2.1, and Qwen2-VL for role-ID selection. Under Oracle, both agent-based methods use the same provided event interval and shared spatial backend. Their Oracle scores therefore primarily reflect the shared spatial grounding pipeline rather than independent spatial grounding capabilities.

\textbf{Evaluation Verification} We use the original Event-STVG evaluator without ground-truth-assisted correction of predicted intervals, role identities, or bounding boxes. Unsuccessful predictions are retained as zero scores.

\textbf{Agent Mechanism Validation} VideoTemp-o3 did not invoke its optional tools during Event-STVG task. To verify that the tool mechanism was functional under the same frozen deployment configuration, we evaluate its native agent on 30 official Video-MME questions using both frozen prompt paths, together with five repeated positive-control runs per path. The Oracle and 1-minute labels refer to the original prompt paths; both use the same full videos.

\begingroup
\setlength{\tabcolsep}{2pt}
\renewcommand{\arraystretch}{1.12}
\setlength{\LTcapwidth}{\linewidth}
\setlength{\frozenevaltablewidth}{\dimexpr\linewidth-8\tabcolsep\relax}
\begin{longtable}{@{}>{\raggedright\arraybackslash}p{0.170\frozenevaltablewidth}>{\raggedright\arraybackslash}p{0.205\frozenevaltablewidth}>{\raggedright\arraybackslash}p{0.215\frozenevaltablewidth}>{\raggedright\arraybackslash}p{0.205\frozenevaltablewidth}>{\raggedright\arraybackslash}p{0.205\frozenevaltablewidth}@{}}
\caption{VideoTemp-o3 tool execution under the frozen deployment configuration.}\label{tab:appendix-videotemp-tools}\\
\toprule
Prompt path & Tool requested & Successful crops & Feedback processed & Answer extracted \\
\midrule
\endfirsthead
\multicolumn{5}{@{}l}{\tablename\ \thetable\ (continued)}\\
\toprule
Prompt path & Tool requested & Successful crops & Feedback processed & Answer extracted \\
\midrule
\endhead
\multicolumn{5}{r@{}}{Continued on next page}\\
\endfoot
\bottomrule
\endlastfoot
Oracle & 28/30 (93.3\%) & 41/45 (91.1\%) & 41/41 (100\%) & 30/30 (100\%) \\
1-minute & 26/30 (86.7\%) & 28/29 (96.6\%) & 28/28 (100\%) & 24/30 (80.0\%) \\
\end{longtable}
\endgroup

\textit{Tool requests are counted per QA, successful crops per tool request, and feedback processing per successfully decoded clip. Answer extraction measures output availability, not correctness.}

Both positive-control paths successfully complete the tool workflow in all five runs. Across all 70 runs, the model generates 88 parsable tool requests, of which 83 produce decodable clips that are subsequently passed back into model inference. The five failures arise from out-of-range crop requests. We also observe playback-duration deviations exceeding 50 ms in 35 of the 83 returned clips, reflecting limitations of the original sampling implementation. In the 1-minute prompt path, none of the 30 outputs satisfies the frozen temporal-window JSON protocol.

These results confirm that VideoTemp-o3 can autonomously invoke and execute video tools under the evaluated configuration.

\clearpage
\subsection{Same-Type Mismatches Among Target Misses}
\label{app:same-type-misses}

\begin{table}[H]
\centering
\caption{Proportion (\%) of target misses overlapping another qualifying same-type occurrence on the fixed 355-query diagnostic subset with sufficient occurrence-level temporal evidence.}
\label{tab:same-type-misses}
\renewcommand{\arraystretch}{1.12}
\begin{tabular*}{\linewidth}{@{\extracolsep{\fill}}lrrrr@{}}
\toprule
Model & 1 min & 3 min & 5 min & 10 min \\
\midrule
Gemini & 20.00 & 24.24 & 27.78 & 38.07 \\
GPT & 28.95 & 40.82 & 46.25 & 58.56 \\
Qwen & 11.90 & 37.97 & 33.90 & 36.81 \\
\bottomrule
\end{tabular*}
\end{table}

\noindent\textit{Note:} For each model and duration, the denominator is the number of target misses within the same 355-query subset. Misses follow the continuous-time overlap criterion used in the same-type mismatch diagnostic. Figure~\ref{fig:diagnostic-decomposition}(d) instead divides mismatch counts by 355; the grid hit rate in Figure~\ref{fig:diagnostic-decomposition}(a) cannot be used to recover the denominators here. For example, GPT has 111 target misses at 10 minutes, of which 65 overlap another qualifying same-type occurrence, yielding $65/111 = 58.56\%$.

\clearpage
\subsection{Oracle Candidate-ID Selection}
\label{app:oracle-id-selection}

\textbf{Objective} To examine semantic-role identification when the target event interval and spatial candidates are supplied, we evaluate models under an Oracle candidate-ID selection setting. Models identify the entities filling the queried roles and select their displayed IDs, removing the need for temporal retrieval and free-form bounding-box generation. We additionally evaluate whether the queried role fillers can be identified jointly at the same timestamp.

\textbf{Sampling} We construct a fixed cohort of 300 queries from the 3,516-event soccer Oracle population. We first retain the 2,904 events whose annotated intervals are strictly longer than two seconds. Event-type quotas are approximately proportional to the eligible category frequencies, using a lower-bounded largest-remainder allocation with at least one query per category.

Within these quotas, selection prioritizes retaining previously accepted tracked-ID inputs, reusing a fixed pool of previously successful and parseable outputs, and preserving overlap with the existing soccer long-context subset, in that order. Seeded shuffling resolves choices within equal-priority groups, with seed 21 used for the final cohort. No grounding scores or candidate-selection accuracy values enter selection. The resulting cohort covers all 17 event categories and 48 matches, comprising 90 single-role, 119 two-role, and 91 three-role queries. This is an event-stratified selection with input-reuse priorities, rather than a simple random sample of the full soccer population.

\textbf{Setup} Each query uses its frozen Oracle-trimmed video, event definition, queried-role definitions, and annotated half-second timestamp sequence. Candidate player detections and tracks are generated using SoccerMaster, supplemented with frozen manual candidates when automatic detections are insufficient. Role-bearing candidates and distractors share the same rendering style, without semantic-role labels or correct-ID markers.

We quantify numeric-ID continuity along the GT role trajectories at the frozen 2-FPS evaluation points. Among 4,933 adjacent pairs in which the same role is visible at both points, 4,482 retain the same ID (90.86\%). Of 611 contiguous visible segments containing at least two sampled points, 414 preserve one ID throughout (67.76\%). Across all visible points of each role trajectory, including reappearances after invisible gaps, 401 of 601 trajectories use a single ID (66.72\%). These statistics describe ID continuity at annotated timestamps rather than identity correctness on every intermediate frame.

Approximately 30\% of the selected clips inherit prior human review of trajectory identity associations, and their reviewed assignments are retained. The review checks whether these associations are sufficiently reliable for role selection while allowing track fragmentation and renumbering; it does not require a perfectly constant ID throughout each clip.

Candidates are rendered using thin yellow boxes and black badges with yellow integer IDs. Role-bearing candidates and distractors use the same visual style, without semantic-role labels, correct-ID markers, or candidate provenance. IDs are initialized from automatic tracks and available reviewed identity mappings. Since tracks may fragment or switch identity, models must follow the actual role-bearing person and report the ID currently displayed on that person.

Gemini-3.7 Flash, GPT-5.6 Sol, and Qwen3.8-Max are evaluated on the same 300 queries. Qwen receives the annotated MP4 through its native-video interface with provider-default inference settings. GPT receives ordered frames sampled at 2 FPS and resized to 512\(\times\)288, with low image detail. Gemini uses the accepted 2-FPS, 512\(\times\)288 frame configuration. Optional generation parameters are left at provider defaults.

For each prescribed timestamp, the model returns an integer candidate ID or explicit null for every queried role. ID Accuracy is the number of correct IDs divided by the number of GT-visible role--frame decisions, pooled across the cohort. GT-invisible decisions are excluded; wrong IDs, null, missing answers, invalid IDs, duplicate timestamps, and failed responses count as incorrect on GT-visible decisions. Joint ID Accuracy requires every GT-visible role at a timestamp to be selected correctly, considering timestamps with at least one visible GT role. Thus, the joint metric includes single-visible-role timestamps.

Correct IDs are established by matching frozen candidate boxes to GT boxes with IoU \(\geq\) 0.99 solely to accommodate coordinate rounding; every GT-visible decision has a unique matching candidate. Prediction correctness is subsequently determined by exact ID equality. We score the last saved response per query, reparsed using the frozen extractor. Confidence intervals, where reported, use match-clustered bootstrap with 20,000 resamples across the 48 matches (seed = 21), recomputing pooled accuracy in each replicate. Shared match resamples preserve pairing between models.

\textbf{Prompt} The following is a verbatim three-role example from the fixed cohort. The numbered video or ordered video frames are supplied as the accompanying visual input. Timestamps are measured relative to the Oracle clip, and the schema contains type placeholders rather than correct candidate IDs.

\textbf{SETTING-SPECIFIC TASK CONTEXT}\par

You are given a manually annotated and temporally trimmed soccer video clip of \textbf{Shot off Target}. The entire clip is the annotated temporal window of the target event instance, from its annotated start to its annotated end.  No temporal localization is required.

\textbf{EVENT DEFINITION}\par

\textbf{Shot off Target}: An attacking sequence culminating in an intentional attempt to score where the resulting shot is off target.

\textbf{ROLE DEFINITIONS}\par

\begin{enumerate}
\renewcommand{\labelenumi}{\arabic{enumi})}
\item \texttt{last\_passer}: The attacking player who makes the final effective pass to the shooter before the shot.
\item \texttt{shooter}: The attacking player who takes the shot.
\item \texttt{goalkeeper}: The goalkeeper defending the goal targeted by the attacking sequence.
\end{enumerate}

\textbf{RULES}\par

\begin{enumerate}
\renewcommand{\labelenumi}{\arabic{enumi})}
\item Track the same player for each role throughout the event. If not reliably visible, output \texttt{null}; resume the same player when visible again.
\item At each of the following timestamps, measured from the beginning of the input video, select the numbered bounding box containing each role-bearing person and output its integer ID: [0.0, 0.5, 1.0, 1.5, 2.0, 2.5, 3.0, 3.5, 4.0]
\item Candidate IDs are assigned using automatic tracking and are intended to remain consistent over time, but tracks may be fragmented or contain identity errors. At each timestamp, identify the same role-bearing person and report the ID currently shown on that person. Change the reported ID when needed to follow that same person; do not rely on ID continuity alone.
\item Use each listed timestamp exactly once and do not add other timestamps.
\end{enumerate}

\textbf{STRICT JSON SCHEMA}\par

Return only the JSON array strictly in the format below:

\begin{verbatim}
[
  {
    "time_sec": number,
    "last_passer": integer or null,
    "shooter": integer or null,
    "goalkeeper": integer or null
  }
]
\end{verbatim}

\noindent\textit{Note:} To account for discontinuities in SoccerMaster-generated IDs, Rule~3 explicitly instructs models to follow the same role-bearing person and report the ID currently displayed on that person, changing the reported ID when necessary rather than relying on ID continuity alone.

\clearpage
\subsection{Human Baseline of the ID Selection Setting}
\label{app:human-id-selection}

\textbf{Objective} To assess task feasibility and provide a human reference for interpreting model performance, we evaluate human semantic-role identification under the Oracle candidate-ID selection setting. The target event interval and candidate boxes are supplied, allowing annotators to focus on identifying role fillers and selecting their corresponding IDs.

\textbf{Setup} We select 30 queries from the frozen 300-query soccer Oracle ID-selection cohort using event-type-stratified sampling (seed = 21). We reserve one query for each of the 17 event categories and allocate the remaining 13 places proportionally to the remaining targets within each category. The subset spans 22 matches and contains 8 single-role, 17 two-role, and 5 three-role queries. This 30-query draw uses event labels and fixed cohort membership without consulting model predictions or scores.

Three annotators evaluate the same 30 queries. Each query provides an Oracle-trimmed clip, event and role definitions, and numbered candidate boxes. At each prescribed half-second timestamp, annotators select the displayed ID for each queried role or explicitly record null when it cannot be reliably localized. Since tracked IDs may fragment or switch, annotators must follow the role-bearing person rather than rely on ID continuity alone.

We report ID Accuracy over 458 GT-visible role--frame decisions, excluding 98 GT-invisible decisions. Selecting the GT-associated candidate ID is correct; wrong IDs, null, missing answers, and invalid outputs are incorrect on these visible decisions. Joint ID Accuracy is evaluated over 263 timestamps containing at least one visible GT role and requires all visible roles at a timestamp to be selected correctly. Both metrics pool correct and eligible counts within each annotator.

\textbf{Results}

\begingroup
\setlength{\tabcolsep}{2pt}
\renewcommand{\arraystretch}{1.12}
\setlength{\LTcapwidth}{\linewidth}
\setlength{\frozenevaltablewidth}{\dimexpr\linewidth-4\tabcolsep\relax}
\begin{longtable}{@{}>{\raggedright\arraybackslash}p{0.3\frozenevaltablewidth}>{\raggedright\arraybackslash}p{0.35\frozenevaltablewidth}>{\raggedright\arraybackslash}p{0.35\frozenevaltablewidth}@{}}
\caption{Human performance (\%) in Oracle candidate-ID selection.}\label{tab:human-id-selection}\\
\toprule
Annotator & ID Accuracy & Joint ID Accuracy \\
\midrule
\endfirsthead
\multicolumn{3}{@{}l}{\tablename\ \thetable\ (continued)}\\
\toprule
Annotator & ID Accuracy & Joint ID Accuracy \\
\midrule
\endhead
\multicolumn{3}{r@{}}{Continued on next page}\\
\endfoot
\bottomrule
\endlastfoot
Human 1 & 96.51 & 92.78 \\
Human 2 & 98.03 & 95.44 \\
Human 3 & 95.85 & 90.49 \\
Human mean & 96.80 & 92.90 \\
\end{longtable}
\endgroup

Means are computed across the three annotators.

\textbf{Analysis} Human annotators achieve consistently high performance, with ID Accuracy ranging from 95.85\% to 98.03\% and Joint ID Accuracy from 90.49\% to 95.44\%. Their mean scores reach 96.80\% and 92.90\%, respectively. On the same subset, the strongest evaluated model, Gemini-3.7 Flash, achieves 69.65\% ID Accuracy and 56.27\% Joint ID Accuracy, leaving human--model gaps of 27.15 and 36.63 percentage points. These results support the feasibility of the supplied-candidate task for humans while showing that identifying and jointly associating event-specific participants remains challenging for current models.

\clearpage
\setcounter{section}{2}
\section{Dataset Construction, Annotation, and Quality Control}
\label{app:dataset-construction}
\renewcommand{\thetable}{C.\arabic{table}}
\renewcommand{\theHtable}{C.\arabic{table}}
\setcounter{table}{0}

This appendix describes the annotation workflow, domain-specific procedures,
and quality-control evidence for the final Event-STVG benchmark. The benchmark
contains 4,036 target events: 3,516 soccer events and 520 events from seven
additional domains. Each target is associated with an event interval, its
required semantic roles, and the spatial trajectory of each role filler.

\subsection{Video Sources and Selection}

\paragraph{SoccerNet matches}
The soccer videos come from 50 matches selected from the 500-match SoccerNet
collection~\citep{deliege2021soccernetv2}. We allocate match quotas proportionally across its six competitions
using the largest-remainder method, and spread selection across the
2014--2015, 2015--2016, and 2016--2017 seasons within each competition.
Selection uses a fixed random seed of 42. The final sample comprises nine
English Premier League matches, 12 La Liga matches, ten Serie A matches,
five Bundesliga matches, four Ligue 1 matches, and ten UEFA Champions League
matches. These 50 matches provide the 3,516 annotated soccer targets.

\paragraph{Other sports broadcasts}
We collect full-game broadcasts of 20 NCAA men's basketball games, 20 NHL ice
hockey games, and ten NFL American football games from YouTube. The basketball
videos are from the March Madness channel and the American football videos
from the NFL channel; the ice hockey recordings include both NHL-channel
and third-party uploads. Basketball and American football broadcasts are
paired with ESPN play-by-play records, while ice hockey uses official NHL
play-by-play records. These records support candidate retrieval and
query-specific occurrence counting. Event candidates are selected with
category quotas while spreading coverage across games; a fixed seed of 42
breaks equivalent selection preferences. The final annotated subsets contain
150 basketball, 150 ice hockey, and 100 American football targets.

\paragraph{Esports and other domains}
The esports subset uses two League of Legends match recordings, Gen.G versus
Dplus KIA and Liiv SANDBOX versus Kwangdong Freecs, contributing 30 Kill
targets. For the remaining domains, we curate one excavator-operation video,
one first-person pizza-making video, and two episodes of The Voice Georgia
blind auditions from YouTube. They contribute 30 engineering, 30 food
preparation, and 30 entertainment targets, respectively. These are curated
source videos rather than a random sample of YouTube uploads.

\subsection{Annotation Overview}

Annotation proceeds in three stages: event-window annotation, semantic-role
trajectory annotation, and context/occurrence verification. Annotators first
identify the intended event and determine its start and end according to the
event-specific definition. They then identify the entities filling the required
roles and annotate their bounding boxes at approximately 2 FPS. Finally, the
target is mapped into each evaluation context, and its occurrence index is
determined within that input video.

A semantic role is defined by its relation to the target event. For example,
the last passer is the player who directly passes to the shooter in the target
attack, rather than an arbitrary player making a nearby pass. The role filler
must remain the same entity across sampled frames. When that entity cannot be
reliably localized, the annotation is recorded as invisible (\texttt{null}).
Annotators do not substitute a nearby entity for an occluded or off-screen role
filler. Optional roles are included only when a reliable filler is available;
partial invisibility does not remove an otherwise valid role from the query.

The final population contains 7,342 event--role pairs and 58,915 non-null
bounding-box annotations. Counts refer to unique target events and are not
multiplied by the number of context durations.

\subsection{Annotator-Facing Guidelines}

\paragraph{Event windows}
Annotators verify the event type and key action before accepting a candidate.
Temporal boundaries follow the event definition rather than a universal
fixed-duration window. In soccer, the target is primarily annotated in the
standard broadcast view; unrelated replays, celebrations, and close-ups are
excluded when applicable. If a broadcast cut prevents continuation to the
normal semantic endpoint, the last valid moment in the standard view is used.
These soccer-specific view rules are not imposed on domains whose definitions
explicitly involve camera transitions, such as The Voice.

\paragraph{Role identity and spatial annotation}
Automatic detections and tracks provide annotation assistance where available.
Annotators select the appropriate entity, inspect the sampled trajectory,
correct identity switches and inaccurate boxes, and manually add missing boxes
when the role filler is visible. A stable track ID does not establish correct
identity, and an ID change does not by itself indicate a change in role filler.
The decision follows the event semantics and visual evidence.

\paragraph{Visibility}
A box is annotated only when the role filler can be identified and localized.
Occluded, off-screen, or otherwise unlocalizable role fillers receive
\texttt{null} at the corresponding sampling point. A role with no usable
spatial evidence throughout the event is omitted from the final query, while
its source annotation remains in the provenance. This differs from retaining
a queried role that is invisible in only some frames.

\paragraph{Occurrence verification}
Occurrences are defined by the query-specific event category and counted within
the actual input video. Review checks include target identity, event
completeness, occurrence membership, duplicate or replay handling, and the
target ordinal. Manual acceptance applies to the reviewed samples; automatic
release checks and sampled human review are reported separately below.

\subsection{Soccer Annotation}

Existing soccer event annotations are used to retrieve candidate occurrences
and associate them with the broadcast timeline. Their point timestamps are
not adopted directly as event intervals. Annotators inspect the surrounding
video and manually determine the event boundaries. Actions with different
semantics are annotated separately, including winning a corner versus taking
the corner, and a foul leading to a free kick versus executing that free kick.
Card-related events refer to the producing foul rather than a later referee
close-up.

Candidate player detections and trajectories are generated using the
SoccerMaster pipeline~\citep{yang2026soccermaster} with DETR, PRTReID, and StrongSORT. Annotators assign
semantic roles and retain control over trajectory selection, correction,
visibility, and manual box completion. These generated candidates are
annotation aids rather than unverified ground truth. For 25-FPS sources,
half-second sample positions follow rounded frame locations such as
$0,13,25,38,50,\ldots$, rather than a fixed 12-frame stride.

The final soccer population contains 3,516 targets. Oracle and context inputs
preserve the target identity while changing the available temporal context;
the context-specific occurrence index is recomputed within each input.
The offside role structure includes the offside passer, offside player, and
\emph{second-last defender}, consistent with the final role specification.

\subsection{Annotation in the Seven Additional Domains}

The final extension contains 520 targets: 150 basketball, 150 ice hockey,
100 American football, and 30 each from esports, engineering, food preparation,
and entertainment. These domains contribute 28 event categories in addition
to the 17 soccer categories. The procedures below describe
the event boundaries and role structures used in these domains.

\paragraph{Basketball}
Fourteen categories distinguish made and missed outcomes for seven shot types:
two-point jump shots, three-point jump shots, layups, hook shots, dunks,
putbacks, and free throws. The event starts when the player attempting to
score gains possession and ends when the shot outcome becomes clear. The
annotated semantic role is the shooter.

\paragraph{Ice hockey}
The four categories are saved shots, goals, missed shots, and blocked shots.
Annotators identify the shooter and the optional last passer. The defending
goalkeeper is included for saved shots, goals, and missed shots; a defending
blocker is used for blocked shots. When the query includes a last passer, the
event starts with that player's puck possession; otherwise it starts with the
shooter's possession. It ends when the shot outcome becomes clear. Individual
queries retain their annotated role subset rather than requiring every role
in the category schema.

\paragraph{American football}
Rushing touchdowns ground the runner, from possession to the touchdown.
Passing touchdowns ground the passer and receiver, from the passer's
possession to the receiver's touchdown. Successful field goals ground the
kicker, from the beginning of the run-up until the kicker leaves the frame.
The event-specific endpoint is retained rather than replaced with a generic
score-display or celebration boundary.

\paragraph{Esports}
League of Legends Kill events ground the killer and victim in the main game
view. The interval begins when the victim first loses health in the combat
leading to the death and ends at the victim's death. Minimap icons, portraits,
kill notifications, and player-camera overlays are not substitutes for the
role fillers in the main view. Multiple kills are ordered by victim death time.

\paragraph{Engineering}
Dumping events ground the excavator bucket. The interval begins when the
bucket starts dumping its load and ends when dumping is complete. The spatial
target is the bucket rather than the entire excavator.

\paragraph{Food preparation}
Four categories distinguish removing dough, placing a pizza onto a peel,
putting a pizza into the oven, and taking a pizza out. Annotators ground the
dough or pizza, rather than the worker. Boundaries follow the specified
transfer: scooping and placing the dough on the table; lifting the pizza and
placing it stably on the peel; lifting the loaded peel and fully transferring
the pizza into the oven; or moving the pizza out of the oven and placing it
in the cardboard box. The relevant separation or stable-placement condition
marks completion.

\paragraph{Entertainment}
The Voice events ground the coach pressing the chair button. The event begins
with the frontal close-up preceding the button press and ends when the camera
cuts away after the press. The endpoint is defined by this annotated camera
transition rather than completion of the chair's physical rotation.

\subsection{Construction of 1-, 3-, 5-, and 10-Minute Contexts}

\paragraph{Target-preserving source crops}
After annotating an event, we construct four context videos of 60, 180, 300,
and 600 seconds by cropping continuous intervals from the same source video.
Each interval contains the complete target event and remains within the
permitted source range. We do not concatenate unrelated clips to create the
longer inputs. The target event, queried roles, sampled boxes, and visibility
annotations are retained across durations. A source timestamp $t$ is expressed
in a context starting at $s$ as $t-s$. Spatial supervision remains restricted
to the target event. Same-type occurrences are recounted within each final
input to generate its Only/First/ordinal query.

\paragraph{Soccer}
For a target interval $[a,b]$ and a source half of duration $L$, a context of
length $D$ starts within
\[
\max(0,b-D)\leq s\leq\min(a,L-D).
\]
The soccer generator samples uniformly from this feasible continuous interval.
For each target and duration, the random seed is the unsigned big-endian integer encoded
by the first eight bytes of
$\operatorname{SHA256}(g\,|\,\texttt{sample\_id}\,|\,\texttt{context\_type})$,
where the fields are serialized in this order with literal pipe separators.
The frozen generation configurations use $g=20260727$ for the 1-, 5-, and
10-minute contexts and $g=42$ for the 3-minute contexts.
The current policy does not require one-minute inputs to have
only one same-type occurrence or longer inputs to contain a fixed number of
distractors. Starts are selected separately for each duration; the resulting
windows need not be nested or centered on the target.

\paragraph{Extension sports}
For basketball, ice hockey, and American football, candidate starts are
enumerated at integer-second positions within the source recording. The
planning procedure requires the target and intersecting same-type intervals
in the complete manual target inventory to remain uncut. It prioritizes
the highest feasible target ordinal under that planning inventory and selects
a start uniformly among the corresponding feasible starts (seed~42).
The released inputs use the verified, frozen crop boundaries. Final counts
and ordinals are determined from the full original source-event catalog,
including occurrences that were not selected as targets, rather than only
from the planning inventory.

\paragraph{Other extension domains}
For engineering, food preparation, and entertainment, we enumerate feasible
integer-second starts that contain the target and do not partially cut an
earlier same-type event intersecting the context. Within each domain, event
type, and duration, the planner assigns feasible target ordinals to minimize
the sum of squared ordinal-bin counts, then samples a start conditional on
the assigned ordinal (seed~42). League of Legends uses separately selected
random contexts with the same target and earlier-occurrence completeness
requirements; it is not retrospectively treated as an ordinal-balanced
sample. Subsequent same-type events cut by the right boundary are excluded
from complete-occurrence counts under the non-sports extension rules.

\paragraph{Query records and unique contexts}
Table~\ref{tab:unique-contexts} summarizes the query-input and unique-context
counts. The four durations each contain 4,036 query-input records, for 16,144
records in total. We identify a unique context by its physical source-timeline ID and
exact crop start and end positions; the two halves of a soccer match are
separate timelines. The 1-, 3-, 5-, and 10-minute settings contain 4,034,
4,032, 4,033, and 4,024 unique contexts, respectively, totaling 16,123.
Thus, deduplication merges 21 records (0.13\% of all query-input records).
A small number of distinct target queries share exactly the same source
interval; they remain separate queries about different target events.
These exact duplicates are distinct from partially overlapping contexts,
whose shared footage is quantified below.

\paragraph{Overlap between context inputs}
We measure shared source footage in two ways. For distinct targets $i\ne j$
from the same physical source timeline and at the same duration, we compute
\[
J_{ij}=\frac{|W_i\cap W_j|}{|W_i\cup W_j|},\qquad
C_{ij}=\frac{|W_i\cap W_j|}{D}.
\]
All unordered pairs are included, including disjoint pairs, and pairs receive
equal weight. Different soccer halves are separate timelines. Each duration
contains 4,036 inputs across 156 source timelines, yielding 66,645 such pairs.
Table~\ref{tab:context-overlap-within} reports the pooled results, and
Table~\ref{tab:context-overlap-domains} gives the domain breakdown.
Averaging equally over the 1-, 3-, 5-, and 10-minute settings, the overall
mean pairwise temporal IoU is 8.88\%.

For the same target across two durations, we report temporal IoU, the fraction
of the shorter window covered by the longer window, and complete nesting
(Table~\ref{tab:context-overlap-cross}). The shorter window is fully contained
in the longer one for 67.81\% of the one-versus-three-minute pairs and 51.71\%
of the five-versus-ten-minute pairs. Longer-duration inputs therefore preserve
the target but do not always contain every frame of the corresponding shorter
input.

\begin{table}[t]
\centering
\caption{Query-input records and unique contexts in the full benchmark.
Contexts are deduplicated by physical source timeline and exact crop boundaries;
soccer halves are separate timelines.}
\label{tab:unique-contexts}
\begin{tabular}{rrr}
\toprule
Minutes & Query-input records & Unique contexts \\
\midrule
1 & 4,036 & 4,034 \\
3 & 4,036 & 4,032 \\
5 & 4,036 & 4,033 \\
10 & 4,036 & 4,024 \\
\midrule
Total & 16,144 & 16,123 \\
\bottomrule
\end{tabular}
\end{table}

\begin{table}[t]
\centering
\caption{Overlap between different target videos at the same duration in the
full benchmark. All 66,645 unordered pairs from the same source timeline are
included at each duration, including pairs with no overlap. Soccer halves
are separate source timelines. Shared content is intersection duration divided
by the common video duration. Values are percentages.}
\label{tab:context-overlap-within}
\begin{tabular}{rrrr}
\toprule
Minutes & Mean temporal IoU & Mean shared content & Pairs with overlap \\
\midrule
1 & 1.75 & 2.27 & 4.51 \\
3 & 5.46 & 6.98 & 13.51 \\
5 & 9.21 & 11.72 & 22.18 \\
10 & 19.11 & 23.90 & 42.78 \\
\bottomrule
\end{tabular}
\end{table}

\begin{table}[t]
\centering
\caption{Mean temporal IoU (\%) between distinct target videos from the same
source timeline, by domain and context duration. All within-source pairs are
included, including pairs with zero overlap; pairs receive equal weight.}
\label{tab:context-overlap-domains}
\begin{tabular}{lrrrr}
\toprule
Domain & 1 min & 3 min & 5 min & 10 min \\
\midrule
Soccer & 1.73 & 5.40 & 9.11 & 18.84 \\
Basketball & 0.00 & 0.45 & 1.11 & 4.15 \\
Ice hockey & 1.40 & 3.53 & 4.77 & 8.38 \\
American football & 0.00 & 0.07 & 1.16 & 3.58 \\
Esports & 4.47 & 9.09 & 14.94 & 39.62 \\
Engineering & 4.01 & 16.89 & 32.37 & 70.94 \\
Food preparation & 4.22 & 14.93 & 23.77 & 45.06 \\
Entertainment & 4.20 & 4.29 & 4.36 & 12.38 \\
\bottomrule
\end{tabular}
\end{table}

\begin{table}[t]
\centering
\caption{Cross-duration overlap for the same 4,036 targets. Shorter coverage is
the fraction of the shorter video shared with the longer video. Fully nested
is the percentage of targets whose shorter window is completely contained in
the longer window. Temporal IoU has an upper bound equal to the shorter-to-longer
duration ratio. All values are percentages.}
\label{tab:context-overlap-cross}
\begin{tabular}{lrrr}
\toprule
Minutes & Mean temporal IoU & Shorter coverage & Fully nested \\
\midrule
1 vs. 3 & 29.45 & 89.91 & 67.81 \\
1 vs. 5 & 18.60 & 93.61 & 79.48 \\
1 vs. 10 & 9.61 & 96.32 & 87.69 \\
3 vs. 5 & 46.48 & 82.01 & 42.59 \\
3 vs. 10 & 26.70 & 90.39 & 67.77 \\
5 vs. 10 & 41.35 & 85.86 & 51.71 \\
\bottomrule
\end{tabular}
\end{table}


\subsection{Quality-Control Protocol}

\paragraph{Structural checks}
Automatic validation checks parseable files, unique sample identifiers,
ordered and in-range temporal intervals, valid sampling indices, positive-area
boxes in the appropriate coordinate system, and consistency among manifests,
source annotations, prompts, and ground truth. Source annotations and the
derived queries are linked through retained provenance. Visibility records
are preserved through coordinate and timeline transformations.

\paragraph{Semantic and visibility review}
During annotation and review, annotators check that the intended event is
present, the interval follows its definition, each entity satisfies its
semantic role, and role identity remains consistent across frames. They also
check that invisible records reflect an inability to localize the role filler,
and correct visible but missing boxes.

\paragraph{Context-specific occurrence counts}
The occurrence sequence is constructed from same-type events in the same
source video, including qualifying source occurrences that were not selected
as benchmark targets. Each duration is counted separately. For a context with
$N$ qualifying occurrences and target index $k$, the query uses
\emph{only} when $N=1$, \emph{first} when $N>1$ and $k=1$, and the appropriate
ordinal when $k>1$. Replays do not create additional physical occurrences.

The extension's frozen counting procedure retains the distinction between
complete manually annotated intervals and source records available only as
timestamps or alignment bounds. Full intervals must be contained in the
context. Count-only records use the documented proxy ranges and saved
boundary-review evidence; these ranges are not presented as independently
annotated event durations. Sports sequences follow source-record order,
which handles stopped clocks, period resets, and clock corrections. League
of Legends uses victim death time; engineering uses event start time; food
and entertainment use their saved event timestamps with stable tie handling.

The final extension audit reconstructs all 2,080 context sequences and checks
their membership, counts, ordinals, and query wording. For the sports domains,
we additionally apply two methods to determine occurrence membership and
target ordinals, followed by human verification, as described below.

\subsection{Annotation Cross-Verification and Final Audits}

\subsubsection{Annotation Cross-Verification}

\paragraph{Objective}
We assess the consistency of event boundaries, semantic role grounding, and
visibility annotations across annotators.

\paragraph{Setup}
We sample 355 events from 3,551 soccer event windows after annotation. After
matching them to the final benchmark, 354 events are retained. We additionally
sample 52 events from the cross-domain extension, yielding 406 reviewed events
(10.06\% of the 4,036 evaluation instances).

All reviews are conducted by annotators who did not participate in the
original annotation of the corresponding events. For temporal verification,
reviewers can see the original interval as a read-only reference but
independently determine the boundaries. For soccer, spatial verification uses
the same SoccerMaster candidate tracks without revealing the original
semantic role assignments. Spatial annotations in the other domains are
independently produced.

\paragraph{Results}
Table~\ref{tab:annotation-consistency} reports the updated consistency results.
The jointly visible box statistic uses role--frame pairs with a box in both
annotations. The frame-level micro statistic uses pairs with a box in at least
one annotation and assigns zero IoU to one-sided missing boxes; shared absence
is excluded from this statistic.

\begin{table}[t]
\centering
\caption{Temporal and spatial consistency under annotation cross-verification
with assisted review.}
\label{tab:annotation-consistency}
\begin{tabular}{p{0.78\linewidth}r}
\toprule
Metric & Result \\
\midrule
Mean temporal IoU & 89.74\% \\
Median temporal IoU & 1.000 \\
Temporal IoU $\geq 0.5$ & 95.32\% \\
Temporal IoU $\geq 0.7$ & 83.25\% \\
Mean start-time difference & 0.211 s \\
Mean end-time difference & 0.427 s \\
Mean box IoU (both visible) & 92.51\% \\
Frame-level micro IoU (missing boxes counted as zero) & 90.16\% \\
\bottomrule
\end{tabular}
\end{table}

\paragraph{Analysis}
The results show high agreement in temporal boundaries, spatial localization,
and visibility judgments under the review protocol. However, because temporal
reviewers can see the original intervals and soccer reviewers share candidate
tracks, these results measure consistency under assisted review rather than
fully independent annotation agreement.

\subsubsection{Soccer Final Review and Consistency Checks}

\paragraph{Final temporal review}
We uniformly sample 176 targets without replacement from the 2,464 soccer
targets whose 1-, 5-, and 10-minute context videos were all available at the
time of selection. Sampling is not stratified by event type. The same target
list is subsequently retained when the 3-minute view is added. All selected
targets belong to the final 3,516-target soccer population.
The review interface presents the Oracle clip, the corresponding original
video interval, and the context videos together, allowing reviewers to inspect
event correspondence and the target's start and end positions. All 176
selected targets pass the final temporal review. The fixed sample list and
saved decisions are retained for reproducibility.

\paragraph{Cross-duration consistency}
Oracle and all four context settings contain the same 3,516 soccer target
IDs, and their manifests pass the recorded window-validity checks.

\paragraph{Spatial and visibility consistency}
Each setting preserves the same active role sets, sampled spatial boxes, and
visibility records. The canonical soccer ground truth contains 6,497 active
event--role pairs, 53,604 non-null boxes, and 7,297 null role--frame records.
Checks of the final ground truth find no invalid boxes or active roles lacking
spatial evidence throughout the event.

\paragraph{Occurrence-record consistency}
Across the four context settings, all 14,064 soccer context records are
checked for unique source-occurrence IDs ordered by canonical source-event
time, agreement between occurrence membership and count, and consistency of
the target's recorded ordinal with its position in the sequence.

\subsubsection{Extension Occurrence-Order Verification and Release Checks}

\paragraph{Two-method verification of sports occurrence order}
The basketball, ice hockey, and American football subsets contain 400 targets,
each evaluated under four context durations, yielding 1,600 context windows.
We determine occurrence membership, same-type counts, and target ordinals
using two methods.

The first method aligns the complete source event records to the source-video
timeline and determines which same-type occurrences fall within each context.
It uses manually annotated intervals where available and retained alignment
bounds and boundary-review evidence for the remaining records. The source
catalog contains 5,688 event records across 50 recordings, including events
not selected as benchmark targets.

The second method uses the period and game-clock readings at the boundaries
of the constructed context video to identify the corresponding span of source
event records. Cases involving stopped clocks, repeated readings, clock
corrections, or uncertain boundary readings are resolved using additional
in-video clock observations, source-video boundaries, and saved case-specific
evidence. Source-record order is preserved when clock readings alone do not
determine the sequence. The two methods produce identical occurrence
sequences, same-type counts, and target ordinals for all 1,600 sports contexts.

\paragraph{Human verification}
We further inspect 160 sports context windows, comprising all 42 priority
boundary cases and 118 stratified random windows (seed~42). Reviewers check
the video to verify the same-type occurrence membership, total count, and
target ordinal. All 160 reviewed windows pass verification, and the reviewed
counts and ordinals agree with the final release. Thus, sports occurrence
order is supported by agreement between the two determination methods and
by an additional human verification of boundary cases and sampled contexts.

\paragraph{Final release consistency}

The final extension contains 520 unique targets with 845 active event--role
pairs, 5,311 non-null boxes, and 650 null role--frame records in the canonical
Oracle ground truth. These counts use each target once. Across Oracle and the
four context durations, the release audit checks 2,600 prompt/ground-truth
pairs and verifies their source annotation links. These pairs comprise
520 Oracle inputs and 2,080 context inputs ($520\times4$). Across all seven
extension domains, the 2,080 context occurrence sequences match the verified
membership, count, ordinal, and query-wording records. This release check
ensures that the verified occurrence information is consistently represented
in the final prompts and ground truth.

\end{document}